\documentclass[10pt,twocolumn,letterpaper]{article}
\usepackage[
singlelinecheck=false 
]{caption}
\usepackage[pagenumbers]{cvpr} 

\usepackage[dvipsnames]{xcolor}

\usepackage{amssymb}
\usepackage{pifont}

\definecolor{darkred}{RGB}{204, 0, 0}
\definecolor{darkgreen}{RGB}{0, 160, 0}

\newcommand{\ccmark}{{\color{darkgreen}\ding{52}}}%
\newcommand{\cxmark}{{\color{darkred}\ding{56}}}%

\definecolor{cvprblue}{rgb}{0.21,0.49,0.74}
\usepackage[pagebackref,breaklinks,colorlinks,citecolor=cvprblue]{hyperref}

\def\paperID{****} 
\def\confName{CVPR}
\def\confYear{2024}

\title{Holoported Characters: Real-time Free-viewpoint Rendering of Humans from Sparse RGB Cameras}

\author{Ashwath Shetty\textsuperscript{1,2} \quad
Marc Habermann\textsuperscript{1,3} \quad
Guoxing Sun\textsuperscript{1}\quad
Diogo Luvizon\textsuperscript{1,3}\
\and
Vladislav Golyanik\textsuperscript{1} \quad
Christian Theobalt\textsuperscript{1,3}\
\and
\small{\textsuperscript{1}
  Max Planck Institute for Informatics, Saarland Informatics Campus \quad}
\small{\textsuperscript{2}
  Saarland University}
\and
  \small{\textsuperscript{3}
  Saarbrücken Research Center for Visual Computing, Interaction and AI}\
  \and
}

\begin{document}
\maketitle
\vspace{-20pt}
\maketitle
\begin{abstract}
We present the first approach to render highly realistic free-viewpoint videos of a human actor in general apparel, from sparse multi-view recording to display, in real-time at an unprecedented 4K resolution.
At inference, our method only requires four camera views of the moving actor and the respective 3D skeletal pose.
It handles actors in wide clothing, and reproduces even fine-scale dynamic detail, e.g. clothing wrinkles, face expressions, and hand gestures.
At training time, our learning-based approach expects dense multi-view video and a rigged static surface scan of the actor.
Our method comprises three main stages. 
Stage 1 is a skeleton-driven neural approach for high-quality capture of the detailed dynamic mesh geometry.
Stage 2 is a novel solution to create a view-dependent texture using four test-time camera views as input. 
Finally, stage 3 comprises a new image-based refinement network rendering the final 4K image given the output from the previous stages.
Our approach establishes a new benchmark for real-time rendering resolution and quality using sparse input camera views, unlocking possibilities for immersive telepresence.
Code and data is available on our \href{https://vcai.mpi-inf.mpg.de/projects/holochar/}{project page}.

\end{abstract}
\begin{flushright}
\begin{figure}
\centering
\includegraphics[width=1.0\linewidth]{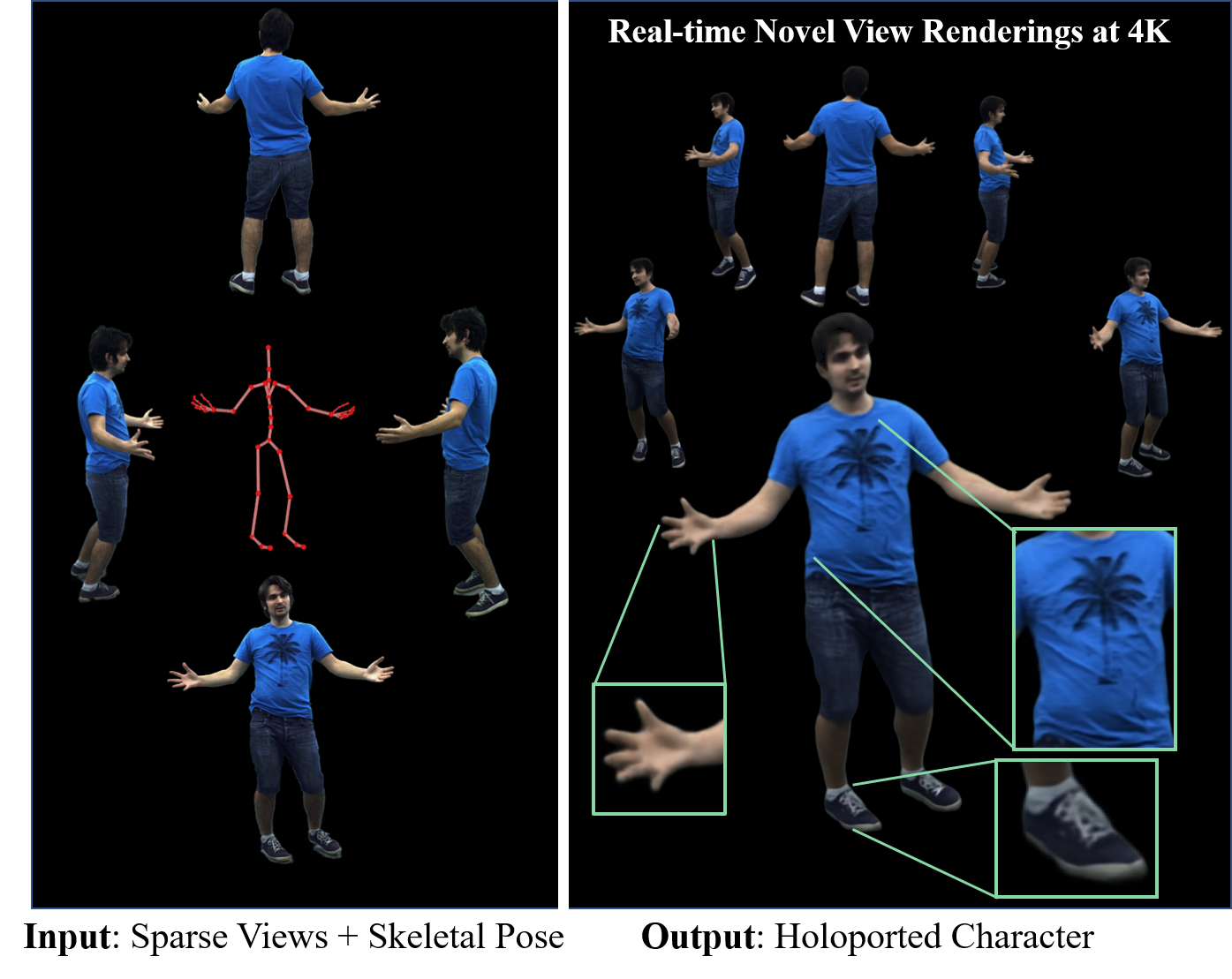}
  \vspace{-12pt}
  \caption{
    We propose \textit{Holoported Characters}, a novel approach for real-time free-view point rendering of humans at 4K resolution. 
    During inference, our method only requires four sparse images observing the human and the respective 3D skeletal pose. 
    Then, our three-stage pipeline generates novel views of the performance in real-time and at an unprecedented resolution of 4K. 
    We highlight that our approach can account for  detailed effects such as clothing wrinkles, facial expressions, and hand gestures.
    }
    \label{fig:1_teaser}
\end{figure}
\end{flushright}
%
%
\vspace{-35pt}
\section{Introduction} \label{sec:introduction}
Human free-viewpoint rendering is a long-standing and highly challenging problem in Vision and Graphics. 
The goal is to render any virtual viewpoint of the character given a discrete set of input camera views.
Earlier approaches resorted to variants of explicit multi-view photometric reconstruction~\cite{tung09,vlasic09,dou16}, reconstruction based on light fields~\cite{mildenhall2019local}, point primitives~\cite{pointbased}, or template-based scene representations~\cite{Xu:SIGGRPAH:2011,casasEG2014,Eisemann08FT} to compute an estimate of dynamic shape and appearance from the multi-view input. 
Despite great progress, these solutions are often limited along multiple dimensions, for example: They often require a very high number of camera views for good quality; computation times are often far from real-time; ghosting artifacts in rendered appearance are prevalent since even the best methods fail to capture error-free scene geometry.
\par
In recent years, a new class of approaches to human free-viewpoint rendering that combines explicit dynamic scene representation with neural-network-based representation and image formation has led to a boost in result quality. These methods utilize neural implicit scene representations to encode the moving human~\cite{habermann2021real,liu2021neural,peng2021animatable,peng2020neural,Shysheya_2019_CVPR, habermann2022hdhumans, Raj_ANR, 2021narf, li2022tava, ARAH:ECCV:2022,remelli2022drivable}. 
However, even the most advanced learning-based approaches face clear limitations: 
Real-time processing from capture to rendering on the basis of neural implicit representations is hard to achieve \cite{habermann2022hdhumans}. 
Rendering resolution is often limited \cite{peng2020neural,lin2022efficient,remelli2022drivable}.
Capturing and displaying fine-scale dynamic effects, such as face expressions, hand gestures, clothing dynamics, or the waving of hair, is often impossible \cite{habermann2021real,liu2021neural,habermann2022hdhumans} or requires complicated tracking and registration pipelines \cite{donglai2022dressing,wang2022hvh}.
Finally, even for some of the best approaches, there are clear differences between ground-truth reference views and renderings, since methods aim for plausibility (where details can differ) rather than truthful detail reproduction \cite{habermann2021real,habermann2022hdhumans, Raj_ANR, 2021narf, li2022tava, ARAH:ECCV:2022} and loose types of apparel are often times out of reach \cite{remelli2022drivable}.
\par 
We, therefore, present Holoported Characters, a new method for human free-viewpoint rendering that is the first combining the following properties: 
It is end-to-end real-time at test time, potentially enabling live capture and free-viewpoint rendering of a human in general wide clothing at unprecedented 4K image resolution. 
It achieves state-of-the-art and truthful, not merely plausible, free-viewpoint rendering quality of even fine details (see Fig.~\ref{fig:1_teaser}). 
It only requires four camera views at test time. 
It truthfully reproduces even face expressions, hand and finger gestures, and dynamic details of loose clothing. 
This combination paves the way for high-quality rendering in real-time merged reality and telepresence. 
\par 
The training phase of our algorithm requires as input a static 3D scan of the person rigged with a skeleton, as well as dense multi-view video of the moving person. 
Our algorithm operates in three stages, each of which introduces important contributions: 
In stage 1, our first contribution is an improved real-time approach for neural network-based skeleton-driven deformation of the template mesh. 
It extends the approach of \citet{habermann2021real} by training the deformation method on the multi-view video using both color-based supervision as well as supervision from neural SDF reconstructions, which greatly enhances 3D shape quality. 
In stage 2, given stage-1 human mesh reconstructions, we propose a real-time projective texturing pipeline that maps the images onto the texture space of the mesh, and a new neural-network-based method computes a dynamic and view-dependent surface texture and feature map from this projected texture. 
It learns to compute a complete coherent surface texture with minimal distortion despite potential inaccuracies in stage-1 geometry.
In stage 3, our new image-based refinement network takes the stage-1 geometry rendered with the stage-2 texture and features as input and transforms it into the final high-resolution rendering.
\par 
We validate our design through thorough ablations, and demonstrate state-of-the-art quality in our experiments.
%
%
%
%
\begin{figure*}
\centering
    \includegraphics[width=0.9\textwidth]{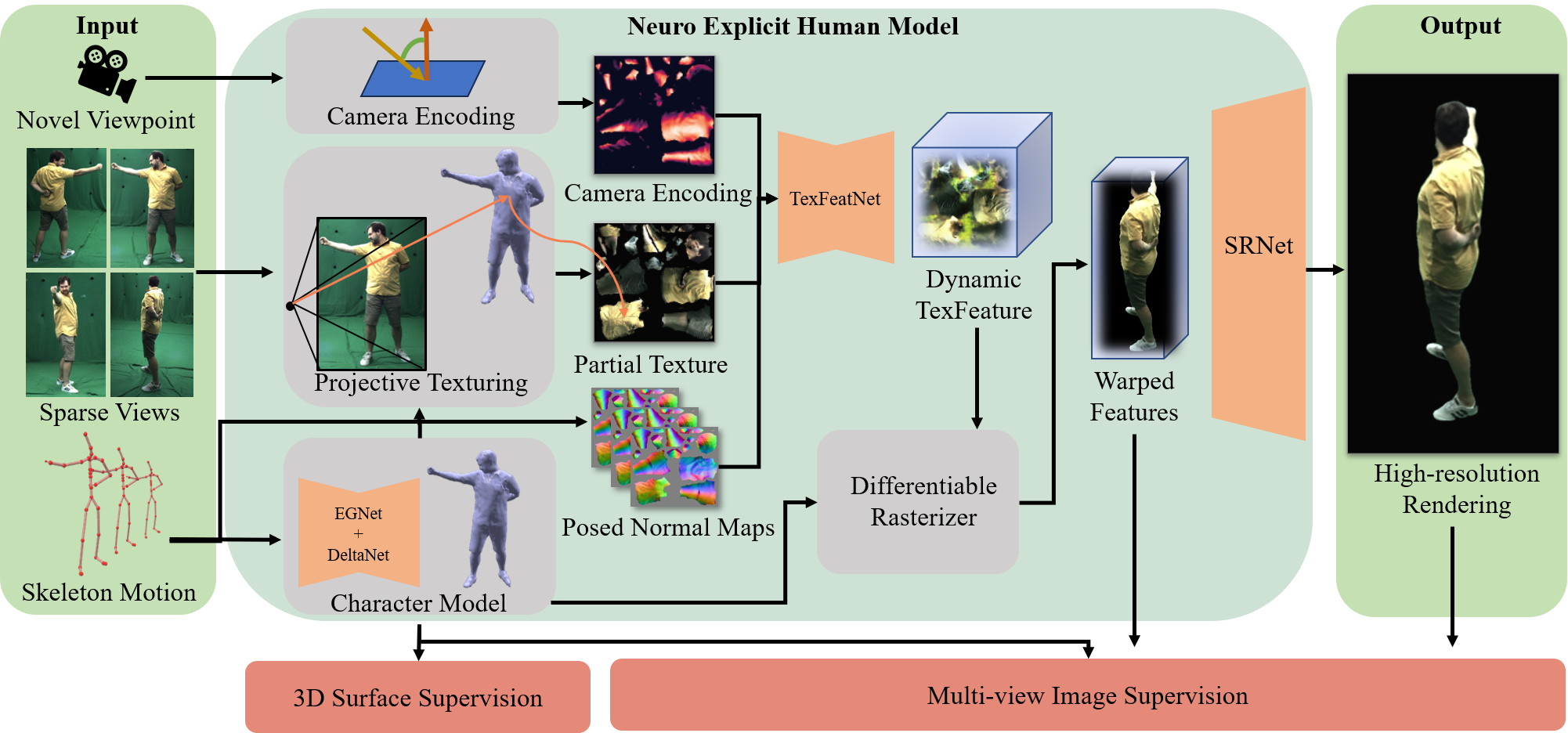}
    \caption{
    \textbf{Method Overview.}
   \textit{Holoported Characters} takes sparse camera views, the respective 3D skeletal pose, and the camera parameters of the novel view as input and generates high-resolution rendering in real-time.
Our character model takes the motion as input and predicts a pose-dependent deformation of the template mesh.
  Then, our projective texturing pipeline maps the sparse views onto this mesh’s texture space.
This texture, camera encoding, and posed normal maps are then fed into our TexFeatNet, producing a view-dependent dynamic texture feature.
  Finally, our SRNet takes those low-resolution features in image space and generates the high-resolution rendering.
    }
  \label{fig:2_pipeline} 
\end{figure*}
\section{Related Work} 

\label{sec:related} 
%
%
\paragraph{Free-view Replay.}
Methods for novel view synthesis of general non-rigid scenes~\cite{tretschk2020nonrigid,Li2022CVPR,liu2023robust,Wang2023FSDNeRF} can be applied to videos with humans but struggle with large articulations due to the absence of human-specific priors. 
They enable scene replay but do not allow changes in the human pose without model retraining. 
Other human-specific replay methods can be trained using monocular \cite{weng2022humannerf} or multi-view videos \cite{HumanRF2023,zhao2022humanp}.
The visual quality of monocular methods can suffer due to the lack of explicit 3D information \cite{weng2022humannerf}. 
Multi-view methods \cite{HumanRF2023,zhao2022humanp} can also render complex appearance elements like hair and clothing, even in real-time \cite{zhao2022humanp}. 
However, as they can only replay the multi-view sequence they were trained on, they are not well suited for interactive telepresence applications, which is the focus of this work. 
%
%
%
\paragraph{Animatable Avatars.}
Recent approaches generate novel 2D views of humans in novel poses but do not allow freely changing the 3D viewpoint \cite{chan2019dance, Siarohin_2019_NeurIPS, kappel2020high-fidelity, liu19, liu20}.
Specialized approaches for the 3D free-viewpoint rendering of humans from monocular or multi-view RGB videos achieve high-fidelity results \cite{weng2022humannerf, xu2021hnerf, liu2021neural, jiang2022neuman, zhang2021stnerf}; some of them generalize to poses unseen during training~\cite{Shysheya_2019_CVPR, habermann2021real, habermann2022hdhumans, Raj_ANR, 2021narf, li2022tava, ARAH:ECCV:2022}. 
Their major limitation is that details such as clothing wrinkles are blurred as the pose information is ambiguous: 
A single pose can induce a variety of wrinkles, which often happens during training and which causes averaging of fine appearance details. 
While HDHumans \cite{habermann2022hdhumans} and other concurrent works~\cite{kwon2023deliffas, zhu2023ash} can generate high-frequency wrinkles (due to the generative formulation), those are often hallucinated and are not consistent with the ground-truth observations. 
Moreover, many methods also strongly rely on a neural rendering component, which makes them slow and not suitable for immersive applications. 
DDC \cite{habermann2021real} is one of the few methods which run in real-time and enable animatable control over the character: It generates an explicit mesh and texture and supports loose clothing. 
However, it still suffers from blurred details as most other methods.
\paragraph{Image-driven Dynamic Scene Rendering.}
Approaches that use images for novel view rendering of dynamic scenes could also be applied to our task \cite{lin2022efficient,yu2020pixelnerf,wang2021ibrnet}. 
The recent work ENeRF~\cite{lin2022efficient} is an interactive and real-time approach for free-viewpoint rendering with a neural representation driven by a sparse set of multi-view images. 
However, due to the lack of human priors, the method suffers from multi-view-consistency artifacts and does not generalize well to new poses and views. 
In contrast, our method achieves view consistent and high-quality results in real-time as we explicitly account for the humanoid structure, i.e. its articulation and non-rigid deformations. 
%
%
\paragraph{Sparse Image-driven Avatars.} 
In contrast to the previous paragraph, some image-driven works~\cite{kwon2023neuralimage,chen2023uv,remelli2022drivable} explicitly model human priors. 
Neural Image-based Avatars~\cite{kwon2023neuralimage} is a generalizable approach, which can drive arbitrary human performers from sparse images and 3D poses. 
However, it fails to generate high-quality wrinkles and expressions for new identities. 
UV Volumes~\cite{chen2023uv} is a recent work that leverages a neural texture stack and generates a UV volume for real-time free-viewpoint rendering. 
However, they require a dense capture setting at test time, whereas our approach requires solely four cameras. 
Drivable Volumetric Avatars (DVA)~\cite{remelli2022drivable} is most closely related to ours in that it achieves high-quality free-viewpoint renderings of humans in real-time from 3D skeletal pose and sparse RGB images. 
In contrast to our method, they represent the virtual character as volumetric primitives loosely attached to a skinned mesh. 
We found that their formulation is limited to tight types of apparel, and their model does not scale well for a large variety of poses. 
Instead, our approach features a deformable character model capable of dealing with loose clothing. 
Moreover, as we optimize explicit structures like meshes and textures, we demonstrate high-quality appearance recovery as well as generalizability to arbitrary poses.
%
%
\section{Method}\label{sec:overview}
Given multi-view images of an actor, our objective is to train a person-specific model capable of generating photoreal renderings during inference. 
The model takes \textit{sparse} multi-view images, the corresponding 3D skeletal pose, and the target virtual camera as input, and produces a rendering for the specified view (see Fig.~\ref{fig:2_pipeline}).
First, we introduce our character model (Sec.~\ref{subsec:ddc}), which generates a posed character mesh from the 3D skeletal pose (stage 1).
Then, we explain our texture projection module (Sec.~\ref{subsec:tex_proj}), which projects the input images onto the mesh and generates a partial texture map effectively encoding the high-frequency visual information of the sparse views (stage 2).  
Following this, our TexFeatNet (Sec.~\ref{subsec:tex_translation}) generates temporally stable, view-conditioned, and complete texture and features for the target view, which are rendered into image space and fed finally to our SRNet module (Sec.~\ref{subsec:super_res}), which generates the final 4K renderings (stage 3).  
%
%
\subsection{Deformable Character Model} \label{subsec:ddc}
First, we introduce our deformable mesh-based model of the human, which is image-agnostic, i.e. only depends on the skeletal motion. 
It will later serve us as a proxy for our projective texturing (Sec.~\ref{subsec:tex_proj}) as we aim at learning the appearance in 2D texture/image space rather than in 3D, as this is efficient and compatible with 2D convolutions.
This model takes the 3D motion $M_t$ as input, and poses a template mesh where $t$ denotes the time. 
We build on top of the learned character model of Habermann et al.~\cite{habermann2021real},
%
\begin{equation}
C_i (M_t, f_{\mathrm{eg}}(M_t), f_{\mathrm{delta}}(M_t)) = \boldsymbol{v}_i,
\end{equation}
%
as it is differentiable, real-time, and has the capacity to effectively model loose clothing.  
The model incorporates a structure-aware graph neural network $f_{\mathrm{eg}}(M_t)$ to address coarse deformations. 
This network predicts the rotation and translation of the nodes of an Embedded Graph \cite{sorkine2007rigid} of the template. 
Simultaneously, the network $f_{\mathrm{delta}}(M_t)$ addresses fine deformations, predicting per-vertex displacements of the template.
\par
\begin{figure}
\centering
    \includegraphics[width=1.\linewidth]{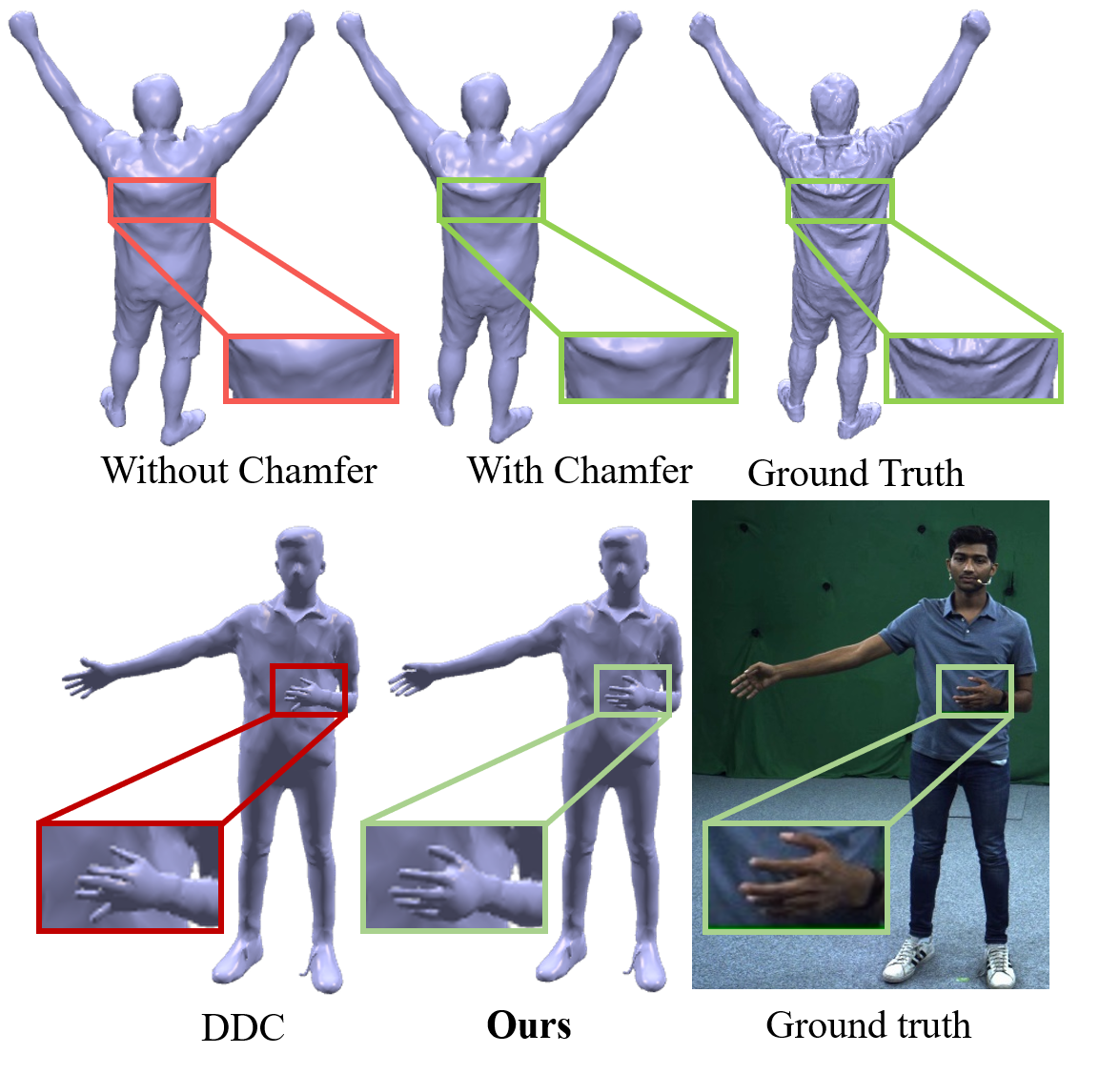}
    \vspace{-24pt}
    \caption{
    \textbf{Recovery of Geometric Details.}
    Our proposed pointcloud supervision and hand modeling helps us recover more details such as wrinkles and hand gestures compared to the baseline. 
    }
    \label{fig:3_chamfer_ablation}
\end{figure}

 
The resultant deformations are applied to the template in canonical space, followed by posing using Dual Quaternion Skinning \cite{dualquaternion2007}, yielding the final positioned vertex $\boldsymbol{v}_i \in \mathbb{R}^{3}$ for each vertex $i$ of the template mesh. 
The aggregate of these posed vertices $\boldsymbol{V} \in \mathbb{R}^{N \times 3}$ is derived by stacking each $\boldsymbol{v}_i$, where $N$ represents the number of vertices in the template.
Their entire model is trained in multiple stages, using only multi-view data.
Please refer to the supplementary material and the original work~\cite{habermann2021real} for more technical details. 
%
%
\paragraph{Improvements to the Character Model.}
In practice, relying solely on multi-view images for model training yields imprecise reconstruction, which hinders our projective texturing pipeline. 
To improve the character model's surface quality, we first reconstruct a high-quality surface $\mathcal{S}$ per frame using recent state-of-the-art surface reconstruction methods \cite{wang2022neus2}. 
We then leverage $\mathcal{S}$ to provide stronger 3D supervision to the displacement network $f_{\mathrm{delta}}(M_t)$.
In practice, we do this with an additional Chamfer loss:
%
\small
\begin{equation}
    \mathcal{L}_\mathrm{cham}(\boldsymbol{V},\boldsymbol{S}) =\sum_{\boldsymbol{v}_{\mathrm{s}} \in \boldsymbol{S}} \min_{\boldsymbol{v} \in \boldsymbol{V}}||\boldsymbol{v}_{\mathrm{s}}-\boldsymbol{v}||_2^2  + \sum_{\boldsymbol{v} \in \boldsymbol{V}} \min_{\boldsymbol{v}_{\mathrm{s}} \in \boldsymbol{S}}||\boldsymbol{v}_{\mathrm{s}}-\boldsymbol{v}||_2^2
    \label{eq:chamfer_loss}
\end{equation}
\normalsize
%
where $\boldsymbol{S} \in \mathbb{R}^{S \times 3}$ is obtained by using Poisson disk sampling \cite{poissondisk2015} on the surface $\mathcal{S}$.
We observe that this term significantly improves the surface quality of the model. 
Fig.~\ref{fig:3_chamfer_ablation} shows the difference in geometry. 
Notice that the wrinkle on the back is significantly better captured. 
%
%
\par 
Additionally, we observe that the coarse deformation, which $f_{\mathrm{eg}}(M_t)$  predicts, causes artifacts for the hands, as they are a part of the body with high articulation very close together, which cannot be modeled by coarse deformation.
To address this issue, we set specific parameters at hand vertices to zero, preventing the model to deviate from the initial skinning deformation. 
This significantly improves the performance, as illustrated in Fig.~\ref{fig:3_chamfer_ablation}.
%
%
\subsection{Efficient Projective Texturing}
\label{subsec:tex_proj}
Now that we have a reasonable 3D proxy from our previous stage, we discuss how to encode the sparse image information, i.e. the four camera views, into a common texture space. 
We choose the texture space as it presents an efficient way of encoding appearance, as it is spatially aligned per frame, and rendering via a generated texture encodes a 3D bias into the network enforcing multi-view consistency. 
This motivates us to generate partial textures from the sparse input images to effectively encode appearance into a 2D representation while still being 3D-aware.
%
%
\par 
For each view $i$, we generate the partial texture $\boldsymbol{T}_{\mathrm{part},i}\in \mathbb{R}^{W \times H \times 3}$ via inverse texture mapping, where $W$ and $H$ denote the texture map dimensions. 
In more detail, each texel $(u,v)$ is transformed into 3D space using the UV mapping of our deformable mesh $\mathbf{V}$ and projected into the input view $i$ where $(x,y)$ denotes the projected image-space coordinates.
We then store the color of the image at $(x,y)$ in the texture space at $(u,v)$.
However, this operation does not account for the fact that some texels might not be visible from a particular view.
%
%
\par 
Thus, we compute per-view visibility masks $\boldsymbol{T}_{\mathrm{vis},i}\in \mathbb{R}^{W \times H}$, which encode if a texel is visible in view $i$.
For a more detailed derivation of the texel visibility, we refer to the supplement.
Moreover, when the 3D surface normal of a texel $(u,v)$ is almost perpendicular to the camera ray $\boldsymbol{d} \in \mathbb{R}^{3}$ of the pixel $(x,y)$, texture projection suffers from distortions.
Thus, despite the visibility, we add another condition encoded as the boolean texture map:
%
\begin{equation}
\boldsymbol{T}_{\mathrm{angle},i}[u,v]=arcos(\boldsymbol{T}_{\mathrm{norm}}[u,v] \cdot \boldsymbol{d})<\delta ,
\end{equation}
%
which ensures that dot product between the texel normal $T_{\mathrm{norm}} \in \mathbb{R}^{W \times H \times 3 }$ and the camera viewing direction is smaller than a threshold angle $\delta$.
Our final computation for whether a texel is valid or not is defined as: 
%
\begin{equation}
    \boldsymbol{T}_{\mathrm{valid},i}=\boldsymbol{T}_{\mathrm{vis},i} \land \boldsymbol{T}_{\mathrm{angle},i},
\end{equation}
%
where "$\land$" is the logical \textit{and} operator.
%
%
\par
\begin{figure}
\centering
  \includegraphics[width=\linewidth]{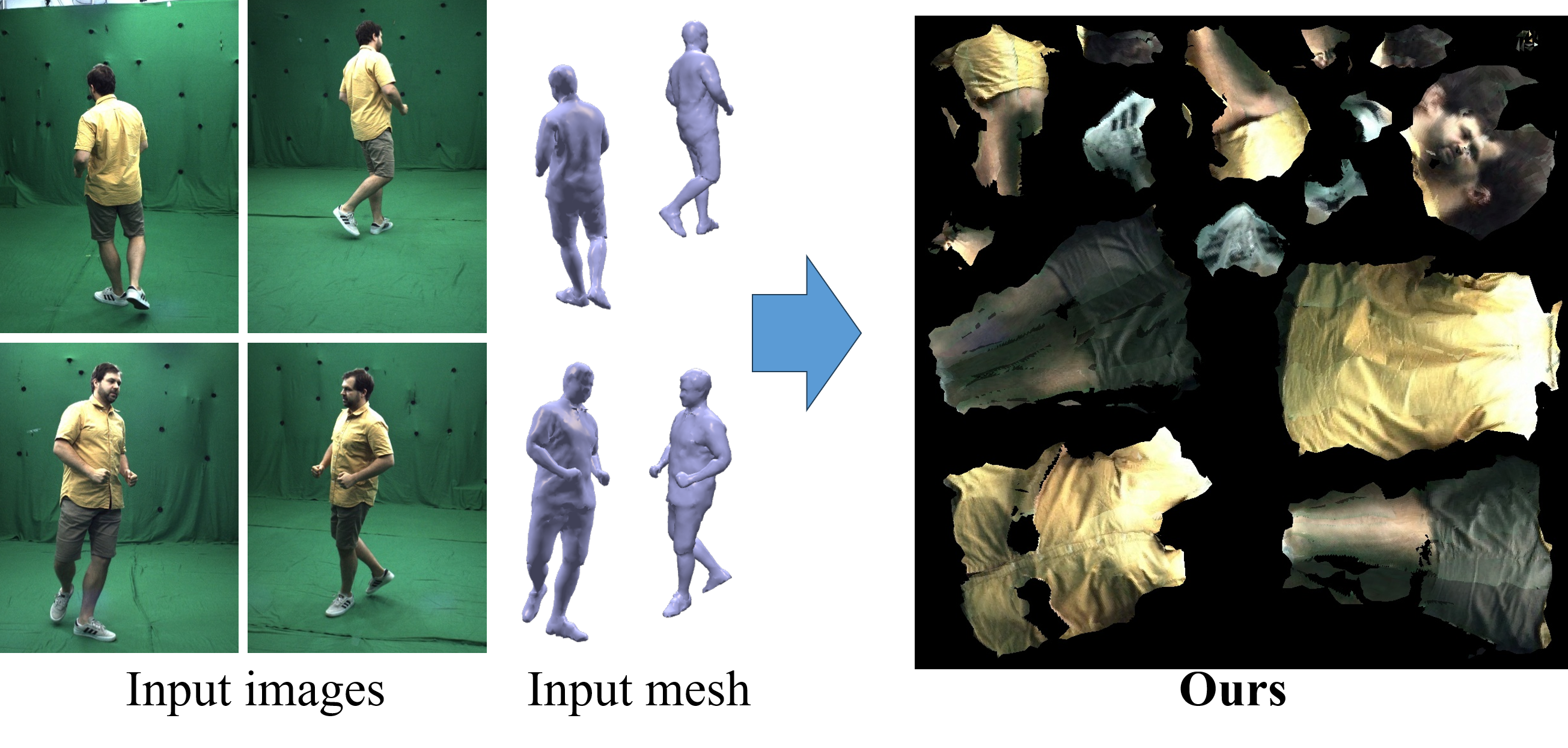}
  \vspace{-12pt}
  \caption{
  \textbf{Projective Texturing.}
  Given a sparse set of cameras and the posed character mesh, our method recovers a partial texture map using projective texturing, where pixels in screen space are mapped to texels of the texture map.
  }
  \label{fig:5_texture_map}
\end{figure}

 
Finally, we fuse the textures from all views via
%
\begin{equation}
    \boldsymbol{T}_{\mathrm{part}}=(\sum_{i \in C_{\mathrm{in}}}\boldsymbol{T}_{\mathrm{part},i} \circ \boldsymbol{T}_{\mathrm{valid},i})\circ(\frac{1}{\boldsymbol{T}_{\mathrm{count}}}),
\end{equation}
%
where "$\circ$" is the Hadamard product, $C_{\mathrm{in}}$ is the number of input camera views, $\boldsymbol{T}_{\mathrm{part}} \in \mathbb{R}^{W \times H \times 3}$ is the final partial texture, and $\boldsymbol{T}_{\mathrm{count}} \in \mathbb{R}^{W \times H}$ stores the number of valid views per texel.
In practice, all operations can be efficiently implemented using tensor operations, thus, leading to real-time computations.
Fig.~\ref{fig:5_texture_map} shows a partial texture obtained by our method.
%
%
\subsection{Tex-2-Tex Translation Network} \label{subsec:tex_translation}
While the generated partial textures effectively encode image information, refinement is essential for achieving photorealistic rendering since imperfect reconstruction and calibration often lead to artifacts. 
Additionally, some details may not be observed from the sparse input views.
Hence, we propose a texture-to-texture network (TexFeatNet):
%
\begin{equation}
f_{\mathrm{tex}}(\boldsymbol{T}_{\mathrm{part}},\boldsymbol{T}_{\mathrm{motion}}, \boldsymbol{T}_{\mathrm{cam}})= \boldsymbol{T}_{\mathrm{dyn}}
= | \boldsymbol{T}_{\mathrm{rgb}}| \boldsymbol{T}_{\mathrm{feat}}| 
\end{equation}
%
\begin{equation}
\boldsymbol{T}_{\mathrm{cam}}=\frac{\boldsymbol{T}_{\mathrm{pos}}-\boldsymbol{o_c}}{||\boldsymbol{T}_{\mathrm{pos}}-\boldsymbol{o_c}||}
\label{eq:camenc}
\end{equation}
%
that takes the sparse image information, geometry as well as viewing direction, and generates a view-dependent and dynamic texture and feature map.
$\boldsymbol{T}_{\mathrm{motion}} \in \mathbb{R}^{W \times H \times T \times 3}$ encodes information about the skeletal motion by baking the 3D surface normals of the posed mesh into a texture and stacking them over a time window of size $T$.
This allows the network to recover pose-dependent appearance features even if such information is not present in the partial texture, i.e. the sparse input views.
$\boldsymbol{o_c} \in \mathbb{R}^{3}$ is the camera origin, $\boldsymbol{T}_{\mathrm{pos}} \in \mathbb{R}^{W \times H \times 3}$ stores the 3D coordinates of each texel and $\boldsymbol{T}_{\mathrm{cam}} \in \mathbb{R}^{W \times H \times 3 }$ encodes the viewing direction of the target view, enabling the network to learn view-dependent effects. 
$\boldsymbol{T}_{\mathrm{part}}$ are the previously derived partial textures.
Without them, we notice that results are blurrier due to the one-to-many mapping~\cite{liu2021neural} between skeletal pose and surface appearance.
\par 
$\boldsymbol{T}_{\mathrm{dyn}} \in \mathbb{R}^{W \times H \times 75 }$ is the output of the network, and contains color ($\boldsymbol{T}_{\mathrm{rgb}}$) in the first three channels and texel features $\boldsymbol{T}_{\mathrm{feat}}$, which are both the input to our super-resolution module introduced next.
%
%
\begin{figure*}
\centering
  \includegraphics[width=0.95\textwidth]{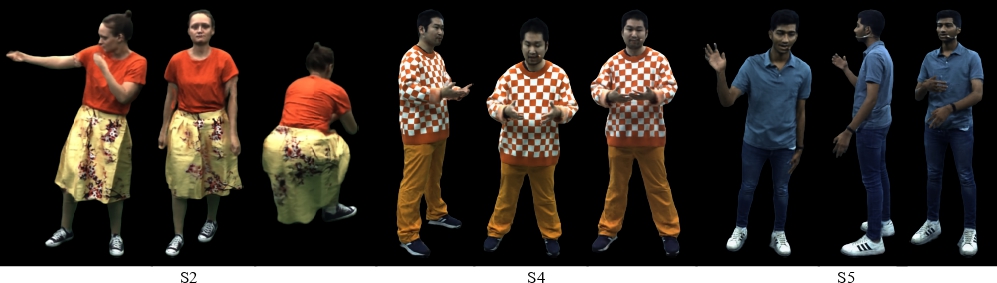}
  \caption{
  \textbf{Qualitative Results for Novel Poses and Views.} 
  Our method generates high-quality renderings showing realistic wrinkle patterns and high-frequency details such as hands gestures and facial expressions.
  Note that our method is robust to challenging poses like squats and complicated clothing types such as loose skirts and highly textured garments, e.g. the pullover.
  }
  \vspace{-10pt}\label{fig:7_novel_pose_single_row}
\end{figure*} 
\subsection{Differentiable Rendering and Super-Resolution} \label{subsec:super_res}
Given the deformable mesh $\mathbf{V}$ (Sec.~\ref{subsec:ddc}) and the first three channels, i.e. $\boldsymbol{T}_{\mathrm{rgb}}$, of the dynamic texture map $\boldsymbol{T}_{\mathrm{dyn}}$, we can already render an image $\mathbf{I}_{\mathrm{mesh},c}$ of the character with dynamic appearance using a standard rasterizer $R$ as 
\begin{equation}
    \mathbf{I}_{\mathrm{mesh},c} = R_c(\mathbf{V}, \boldsymbol{T}_{\mathrm{rgb}}).
\end{equation}
Here, $c$ denotes the target camera view.
However, this suffers from typical mesh rendering artifacts such as staircasing artifacts on the borders and also is limited by the resolution of the texture map.
\par 
To overcome this, we propose a super-resolution module (SRNet), which takes as input $\boldsymbol{I}_{\mathrm{feat},c}$ computed as 
\begin{equation}
    \mathbf{I}_{\mathrm{feat},c} = R_c(\mathbf{V}, \boldsymbol{T}_{\mathrm{dyn}}),
\end{equation}
i.e. by rendering the features and RGB texture onto screen space with a resolution of $W' \times H'$, and outputs a super-resolved image
%
\begin{equation}
f_{\mathrm{sr}}(\boldsymbol{I}_{\mathrm{feat},c})= \boldsymbol{I}_{\mathrm{sr},c},
\end{equation}
%
which has a resolution of $4W'\times4H'$.
As the SRNet purely operates in image space, we implement it as a shallow 2D CNN with small receptive fields.
Thus, it is powerful enough to super-resolve the image with detail recovery while maintaining multi-view consistency. 
%
%
\paragraph{Supervision.}
TexFeatNet $f_{\mathrm{tex}}$ (Sec.~\ref{subsec:tex_translation}) and the SRNet $f_{\mathrm{sr}}$ (Sec.~\ref{subsec:super_res}) are jointly trained. 
The TexFeatNet network is supervised using the rendering loss
%
\small
\begin{equation}
   \mathcal{L}_\mathrm{ren} = \sum_{c,(x,y) \in  \mathbb{R}^{2}} \lVert \boldsymbol{F}_{c}[x,y]\circ (\mathbf{I}_{\mathrm{mesh},c}[x,y]-\boldsymbol{I}^l_{c}[x,y]) \rVert_1,
\end{equation}
\normalsize
%
whereas SRNet is supervised using
%
\small
\begin{equation}
    \mathcal{L}_\mathrm{sr} = \sum_{c,(x,y) \in  \mathbb{R}^{2}}
    \lVert 
    \mathbf{I}_{\mathrm{sr},c}[x,y] -    (\boldsymbol{F}_{c}[x,y]\circ \boldsymbol{I}^h_{\mathrm{c}}[x,y])
    \rVert_1.
\end{equation}
\normalsize
%
$\boldsymbol{F}_{c}$ are the foreground matting masks and $\boldsymbol{I}^l_{c}$ and $\boldsymbol{I}^h_{c}$ are the low- and high-resolution ground-truth images, respectively. 
In $\mathcal{L}_\mathrm{sr}$, we omit applying the matting mask to the super-resolved image, prompting the model to learn complete images with improved border details. 
%
%
%
\section{Results} \label{sec:experiments}
%
%
\paragraph{Runtime.}
Utilizing dual NVIDIA A100 GPUs, our inference pipeline achieves real-time rendering at the full $4112\times{3008}$ resolution. GPU 1 runs the character model and projective texturing at 22 FPS, while GPU 2 handles TexFeatNet and SRNet at approximately 25 FPS. 
See the supplementary material for more details. 
%
%
\paragraph{Dataset.}
We evaluate our method on two sequences from the \textit{DynaCap} dataset~\cite{habermann2021real}, including one subject wearing tight clothes and one subject in loose clothes. 
Since in DynaCap the actors are with fists clenched, we also recorded a new dataset with three novel sequences where the actors were instructed to use both hands in a natural manner; therefore, we can evaluate the capability of our method to represent fingers.
We captured our new sequences in a multi-view camera setup with 120 synced 4K resolution cameras. Each sequence is performed by an actor wearing different clothes and is split into around 20K frames for training and 7K frames for testing. The camera streams are further divided into training and testing views.
To ensure a fair comparison to other methods and to avoid a bias in the results due motion tracking errors (which is not the focus of this work), we run markerless motion capture on 20 cameras.  
%
%
\subsection{Qualitative Results} \label{sec:qualitative}
We evaluate our method qualitatively considering \textit{novel view synthesis} and \textit{novel pose synthesis}. 
Our method generalizes to novel poses at test time as demonstrated in Fig.~\ref{fig:7_novel_pose_single_row}, where we show results for unseen poses rendered under new viewpoints. 
Note the quality of hands, clothing wrinkles, and expressions that can be generated by our method in real time, even for unseen poses. 
We would also like to point out that our approach is able to model wrinkles faithfully, even for very challenging loose clothes, as shown for subject S2. 
This is due to the ability of our character model to generalize in this setting. 
Finally, note the quality of hands and facial expressions, for subjects S4 and S5, which is pivotal for immersive telepresence. 
Additionally, we provide more visualizations and applications such as texture editing as well as telepresence in the supplemental material.
%
%
\subsection{Comparison} \label{sec:comparisons}
%
%
\paragraph{Comparison to Animatible Methods.}
\begin{figure}
\centering
  \includegraphics[width=0.46\textwidth]{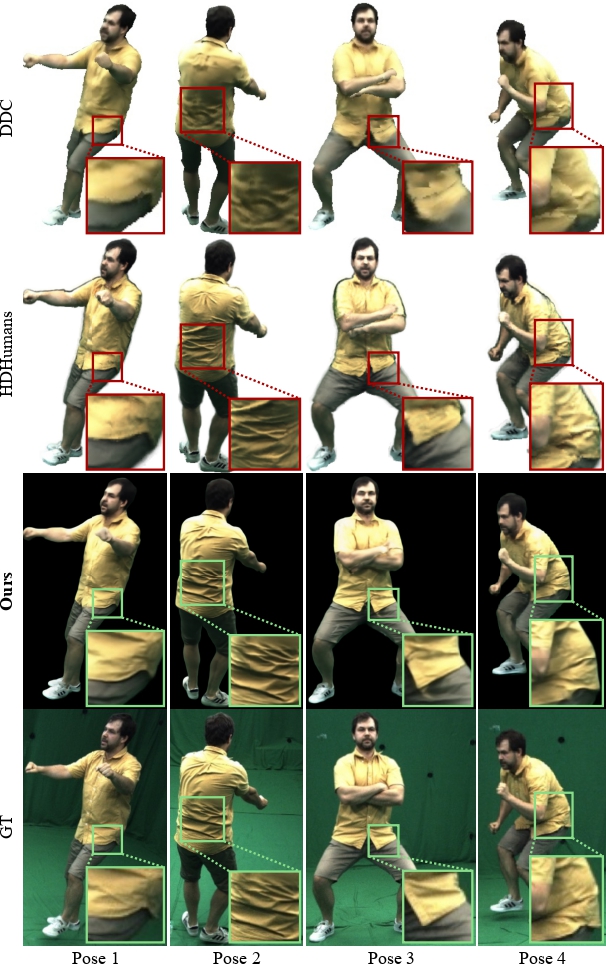}
  \caption{
  \textbf{Comparison with Animatible Approaches.} 
  Here, we show results for novel poses.
  Note that DDC~\cite{habermann2021real} fails to produce high-frequency details and cloth wrinkles.
  HDHumans~\cite{habermann2022hdhumans} can hallucinate high-frequency details, however, they do not match the ground truth. 
  In contrast, our method produces high-quality renderings, which are consistent with the ground truth.
  }
  \vspace{-5pt}
  \label{fig:8_comparison_pose_half}
\end{figure}
In Fig.~\ref{fig:8_comparison_pose_half}, we qualitatively compare our method to approaches that solely take the skeletal pose as input during test time.
DDC~\cite{habermann2021real} is the state of the art in terms of runtime as it demonstrates real-time performance, while HDHumans~\cite{habermann2022hdhumans} can be considered the state of the art in terms of photorealism though not achieving real-time performance.
Note that DDC often synthesizes blurred textures resulting in a lack of detail as the pose to appearance mapping is not a bijection and they do not explicitly account for that.
While HDHumans generates sharper wrinkles, they do not match the ground truth caused by their adversarial training (see insets in Fig.~\ref{fig:8_comparison_pose_half}).
Moreover, both methods cannot account for facial expressions and hand gestures.
In contrast, our method faithfully recovers the ground truth details, and is able to generate facial expressions and hand gestures.
%
\vspace{-5pt}
\paragraph{Comparison to Image-driven Methods.}
\begin{figure}
\centering
  \includegraphics[width=0.47\textwidth]{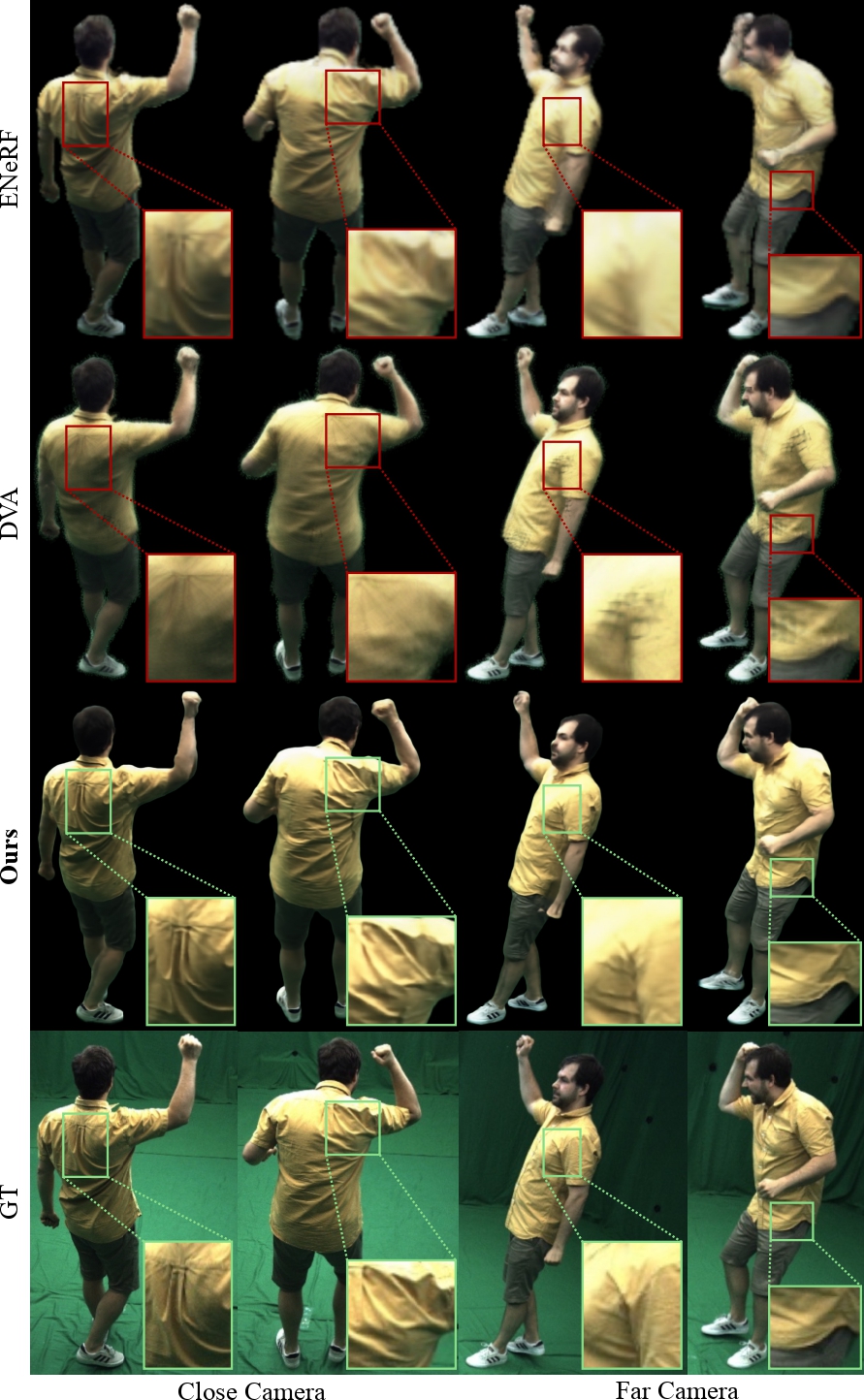}
\vspace{-5pt}
  \caption{
  \textbf{Comparison with Image-driven Approaches.}
  ENeRF~\cite{lin2022efficient} produces artifacts under far views and DVA~\cite{remelli2022drivable} suffers from  blurred renderings. 
 In contrast, our method generalizes well to far views and produces sharp results.
  }
  \vspace{-13pt}\label{fig:9_comparison_enerf_np}
\end{figure}
Next, we compare to methods, that take sparse multi-view images as input and render free-viewpoint videos in real time.
ENeRF~\cite{lin2022efficient} is a method designed for general scenes, while DVA~\cite{remelli2022drivable} is specifically designed for humans, thus, most closely related to our approach. 
Originally, ENeRF uses the \textit{nearest} two views \textit{at test time} from a dense setup to produce the target view, which effectively violates our sparse view requirement during inference.
Instead, we retrain ENeRF using all the training views/frames and always perform inference using four \textit{fixed cameras}, which is the same setting used for all methods.
As shown in Fig.~\ref{fig:9_comparison_enerf_np}, ENeRF, because of their constrained color formulation, fails to maintain multi-view consistency and produces blurry results for novel views, especially for those that are far away from the input views. 
We further compare to DVA~\cite{remelli2022drivable}. 
Since DVA's character model is solely based on skinning and regularizes large deformations, it performs poorly in representing high-frequency details and loose types of apparel. 
We also found that their volume-primitive-based formulation is hard to train on a large variety of poses leading to a degradation in quality when training on large-scale data (such as the DynaCap dataset), which is essential to achieve desired pose generalization at test time.
In consequence, we found that DVA produces blurred results and noticeable artifacts (see insets in Fig.~\ref{fig:9_comparison_enerf_np}).
In contrast, our method results in high-quality renderings with sharp details, wrinkles, and it can also handle loose clothing.
We refer to the supplemental for a more detailed analysis of these methods.
%
%
%
%
\begin{table*}
\small
\begin{center}
\begin{tabular}{@{}cc|ccc|ccc@{}}
    \toprule
     &  & \multicolumn{3}{c|}{Subject S1 (tight clothing)} & \multicolumn{3}{c}{Subject S2 (loose clothing)} \\
     \midrule
                                \textbf{Method}    &\textbf{RT} & \textbf{PSNR}$\uparrow$  & \textbf{LPIPS}$\downarrow$ \footnotesize{($\times 1000$)} & \textbf{FID}$\downarrow$ & \textbf{PSNR}$\uparrow$  & \textbf{LPIPS}$\downarrow$ \footnotesize{($\times 1000$)} & \textbf{FID}$\downarrow$  \\
    \midrule
    DDC	                   & \ccmark        &    32.96 (28.05)          &    20.07 (30.43)             &    27.73 (38.37)     &    27.92 (25.92) &    36.33 (41.07) &    47.23 (56.43)     \\ 
    ENeRF	             &  \ccmark     &    30.75 (30.54)          &    28.03 (29.41)           &     32.81 (36.39)     &    29.83 (\textbf{29.61})  &    35.07 (36.08) &    36.51 (41.08)  \\
     DVA	               &  \ccmark   &    31.65  (\textbf{30.60})        &    26.27  (29.41)           &     35.68 (43.11)
     &    27.32 (24.06) &    40.95 (45.73) &    144.21 (142.07)\\
    \textbf{Ours 1K}	   & \ccmark    &   \textbf{33.93} (30.19)               &  \textbf{14.38} (\textbf{24.95})       &   \textbf{8.27} (\textbf{12.80})     &    \textbf{33.18} (28.85 ) &    \textbf{23.04} (\textbf{30.50})&    \textbf{11.10} (\textbf{18.01})     \\ 
    \midrule
    \midrule
     HDHumans	          &     \cxmark     &    31.00 (27.69)         &    14.61 (24.00)            &     \textbf{4.93} (\textbf{9.25})     &    28.98 (26.32) &    26.28 (33.33) &    6.83 (\textbf{11.87})     \\
    \textbf{Ours}	   & \ccmark    &   \textbf{33.93} (\textbf{30.24})               &  \textbf{12.98} (\textbf{23.74})          &   6.28 (11.62)         &    \textbf{31.45} (\textbf{28.03})    &    \textbf{21.86} (\textbf{28.49})    &   \textbf{5.95} (13.26)        \\ 
    \bottomrule
\end{tabular}
 \vspace{-8pt}
\caption
{
\textbf{Quantitative Comparison.} 
We present \textit{novel view} and \textit{novel pose} (brackets) synthesis results on subjects \textit{S1} and \textit{S2} from the DynaCap dataset~\cite{habermann2021real}. Our method outperforms real-time approaches and particularly excels in loose clothing (\textit{S2}). Bold indicates \textbf{best}. 
}
\label{tab:comparison_nv}
 \vspace{-12pt}
\end{center}
\end{table*}
%
%
%
\vspace{-5pt}
\paragraph{Evaluation Protocol.}
We quantitatively compare our method with the state-of-the-art approaches on two subjects of the DynaCap dataset. 
We trained all methods on every frame of the training sequence using the training camera views.
For evaluating the novel view synthesis accuracy, we compute metrics on holdout views (cameras: 7, 18, 27, 40) on every 10th frame of the training sequence and report the average. 
For novel poses, we compute metrics on the same views but on the testing sequence for every 10th frame.
Again, we report the average across frames.
We report the Peak Signal-to-Noise Ratio (PSNR), Learned Perceptual Image Patch Similarity (LPIPS)~\cite{zhang2018unreasonable}, and Frechet Inception Distance (FID)~\cite{heusel17} in the following.
For a fair comparison, we trained all approaches on 1K resolution. 
%
%
\vspace{-5pt}
\paragraph{Quantitative Comparisons.} 
In Tab.~\ref{tab:comparison_nv}, we report the quantitative comparisons for both tasks, i.e. novel view and pose synthesis.
Our method outperforms all previous real-time methods by a significant margin, especially in LPIPS and FID metrics. 
This demonstrates its strong superiority in producing realistic renderings with fine details.
Comparatively, our method performs on par with HDHumans, while running orders of magnitude faster.
Most importantly, our method recovers the real details present in the ground truth, while HDHumans produces details that are perceptually plausible, but which not necessarily align with the ground truth. This is further verified by the difference in the PSNR metric between our methods.
%
%
\subsection{Ablation} \label{sec:ablation}
%
%
%
\begin{table}
\centering
\small
\begin{tabular}{@{}ccccc@{}}
    \toprule
    \textbf{Method} & \textbf{PSNR}$\uparrow$  & \textbf{LPIPS}$\downarrow$ \footnotesize{($\times 1000$)} & \textbf{FID}$\downarrow$ & \textbf{Res.} \\
    \midrule
    w/o Texture    &   26.44                &  44.34         &   66.13     &   1K       \\ 
    w/o Features	 &   29.89                &  25.52         &   14.02      &   1K       \\ 
    w/o Chamfer	     &   30.04              &  27.47        &   13.81         &   1K    \\
    w/o SR	         &   29.76                 &  27.82         &   15.17       &   1K       \\ 
    w/o 4K Train	 &   30.19               &  24.95         &   12.80          &   1K    \\  
    \textbf{Ours}	 &   \textbf{30.24}     &  \textbf{23.74}   &   \textbf{11.62}          &   1K    \\ 
    \midrule
    w/o 4K Train	 &   \textbf{28.81}   &  33.35         &   18.69          &   4K    \\ 
    \textbf{Ours}	 &   28.75                &  \textbf{32.4}   &   \textbf{17.42}        &   4K     \\ 
    \bottomrule
\end{tabular}
\caption{
\textbf{Quantitative Ablations.}
We evaluate the main components of our method considering \textit{novel pose synthesis} on subject \textit{S1}.
Every component in our pipeline contributes to our final results. Notably, the texture input, the super resolution (SR) module, and the Chamfer loss are the most relevant components.
}
\label{tab:ablation2}	
\end{table}
%
%
\begin{figure}
\centering
  \includegraphics[width=1.\linewidth]{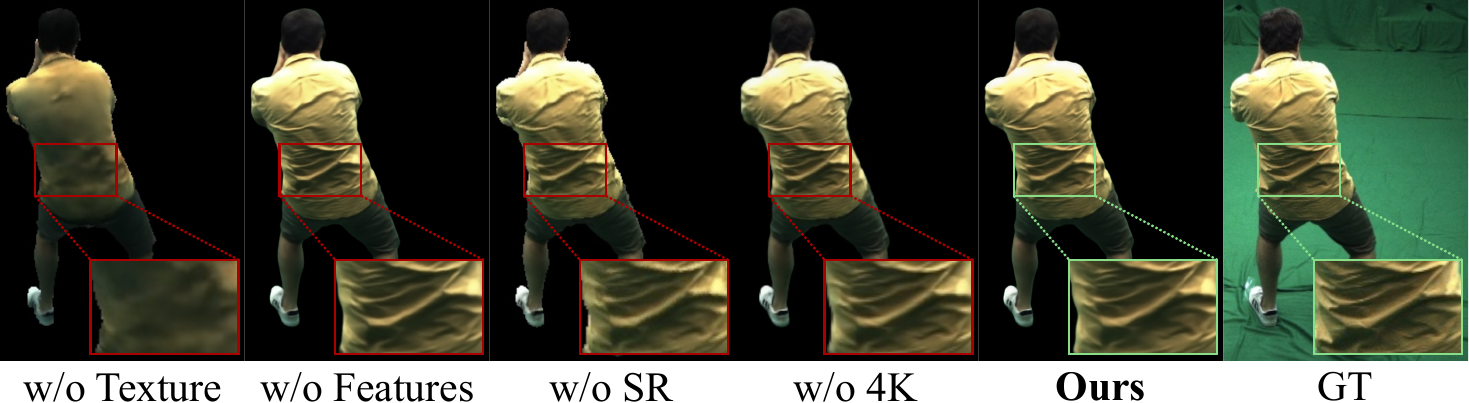}
   \vspace{-8pt}
  \caption{
  \textbf{Qualitative Ablations.}
  Note that all our design choices, i.e. partial texture features, super resolution, and 4K supervision, contribute to the final result quality.
  }
  \vspace{-12pt}
  \label{fig:11_ablation_half}
\end{figure}
%
%
\vspace{-5pt}
\paragraph{Partial Texture Input.}
Removing the partial textures as an input to our TexFeatNet turns our method into an animatable representation, which suffers from the ambiguous mapping from skeletal pose to appearance. 
Adding the partial texture instead effectively provides additional cues and, thus, results in better PSNR, LPIPS, and FID scores (Tab.~\ref{tab:ablation2}). 
We additionally ablate the usefulness of the additional features, which are the output of our TexFeatNet along with the texture itself or, conversely, the input to the SRNet. 
The additional texture features boost all the metrics and also the general sharpness of the result (see ``w/o Features'' and ``Ours'' in Fig.~\ref{fig:11_ablation_half}). 
The additional features help to preserve some of the high-frequency information that might be lost in a pure color-based texture with finite spatial resolution, as well as the forward texture mapping process. 
%
%
\vspace{-5pt}
\paragraph{Geometric Details.}
We also evaluate the effect of the Chamfer loss for the 3D surface supervision in \cref{tab:ablation2}.
It significantly improves the rendering quality, as it leads to surfaces that are better aligned with the input images. 
%
%
\paragraph{4K Resolution.}
The super-resolution (SR) module significantly contributes to improving the visual quality and the metrics.
Importantly, the SRNet removes observable noise, which is due to the discrete mapping of texels, and the typical mesh boundary artifacts (see the inset ``w/o SR'' and ``Ours'' in Fig.~\ref{fig:11_ablation_half}).
Considering the rendering resolution, we observe that training our approach on 4K images translates into better metrics for both evaluation cases (1K and 4K image resolution), especially when considering LPIPS and FID, which are more sensitive to high-frequency details in comparison to PSNR. 
%
%
\section{Conclusions} \label{sec:conclusions}
We introduced Holoported Characters, a novel method for real-time free-viewpoint rendering of humans from sparse RGB cameras and 3D skeletal poses. 
Our approach achieves unprecedented 4K resolution and runtime performance by seamlessly integrating neural and explicit components. 
We believe our work is an important step towards telepresence enabling immersive communication across the globe. However, our method is not without limitations. For example, we cannot handle topological changes such as opening a jacket. For the future, we plan to explore multi-layered human representations potentially being able to model topological changes, which our method currently cannot handle.
\vspace{-5pt}
\paragraph{Acknowledgments.} \label{sec:ack}
This project was supported by the Saarbrücken Research Center for Visual Computing, Interaction and AI. Christian Theobalt was supported by ERC Consolidator Grant 4DReply (770784). 

\clearpage
{
    \small
    \bibliographystyle{ieeenat_fullname}
    \bibliography{references}

\begin{thebibliography}{61}
\providecommand{\natexlab}[1]{#1}
\providecommand{\url}[1]{\texttt{#1}}
\expandafter\ifx\csname urlstyle\endcsname\relax
  \providecommand{\doi}[1]{doi: #1}\else
  \providecommand{\doi}{doi: \begingroup \urlstyle{rm}\Url}\fi

\bibitem[Abadi et~al.(2015)Abadi, Agarwal, Barham, Brevdo, Chen, Citro, Corrado, Davis, Dean, Devin, Ghemawat, Goodfellow, Harp, Irving, Isard, Jia, Jozefowicz, Kaiser, Kudlur, Levenberg, Man\'{e}, Monga, Moore, Murray, Olah, Schuster, Shlens, Steiner, Sutskever, Talwar, Tucker, Vanhoucke, Vasudevan, Vi\'{e}gas, Vinyals, Warden, Wattenberg, Wicke, Yu, and Zheng]{tensorflow}
Mart\'{\i}n Abadi, Ashish Agarwal, Paul Barham, Eugene Brevdo, Zhifeng Chen, Craig Citro, Greg~S. Corrado, Andy Davis, Jeffrey Dean, Matthieu Devin, Sanjay Ghemawat, Ian Goodfellow, Andrew Harp, Geoffrey Irving, Michael Isard, Yangqing Jia, Rafal Jozefowicz, Lukasz Kaiser, Manjunath Kudlur, Josh Levenberg, Dan Man\'{e}, Rajat Monga, Sherry Moore, Derek Murray, Chris Olah, Mike Schuster, Jonathon Shlens, Benoit Steiner, Ilya Sutskever, Kunal Talwar, Paul Tucker, Vincent Vanhoucke, Vijay Vasudevan, Fernanda Vi\'{e}gas, Oriol Vinyals, Pete Warden, Martin Wattenberg, Martin Wicke, Yuan Yu, and Xiaoqiang Zheng.
\newblock {TensorFlow}: Large-scale machine learning on heterogeneous systems, 2015.
\newblock Software available from tensorflow.org.

\bibitem[Casas et~al.(2014)Casas, Volino, Collomosse, and Hilton]{casasEG2014}
Dan Casas, Marco Volino, John Collomosse, and Adrian Hilton.
\newblock {4D Video Textures for Interactive Character Appearance}.
\newblock \emph{Computer Graphics Forum (Proceedings of EUROGRAPHICS)}, 33\penalty0 (2):\penalty0 371--380, 2014.

\bibitem[Chan et~al.(2019)Chan, Ginosar, Zhou, and Efros]{chan2019dance}
Caroline Chan, Shiry Ginosar, Tinghui Zhou, and Alexei~A Efros.
\newblock Everybody dance now.
\newblock \emph{IEEE International Conference on Computer Vision (ICCV)}, 1:\penalty0 0--0, 2019.

\bibitem[Chen et~al.(2023)Chen, Wang, Chen, Zhang, Li, Guo, Wang, and Wang]{chen2023uv}
Yue Chen, Xuan Wang, Xingyu Chen, Qi Zhang, Xiaoyu Li, Yu Guo, Jue Wang, and Fei Wang.
\newblock Uv volumes for real-time rendering of editable free-view human performance.
\newblock In \emph{Proceedings of the IEEE/CVF Conference on Computer Vision and Pattern Recognition}, pages 16621--16631, 2023.

\bibitem[Dou et~al.(2016)Dou, Khamis, Degtyarev, Davidson, Fanello, Kowdle, Escolano, Rhemann, Kim, Taylor, et~al.]{dou16}
Mingsong Dou, Sameh Khamis, Yury Degtyarev, Philip Davidson, Sean~Ryan Fanello, Adarsh Kowdle, Sergio~Orts Escolano, Christoph Rhemann, David Kim, Jonathan Taylor, et~al.
\newblock Fusion4d: Real-time performance capture of challenging scenes.
\newblock \emph{ACM Transactions on Graphics (TOG)}, 35\penalty0 (4):\penalty0 114, 2016.

\bibitem[Eisemann et~al.(2008)Eisemann, De~Decker, Magnor, Bekaert, de~Aguiar, Ahmed, Theobalt, and Sellent]{Eisemann08FT}
Martin Eisemann, Bert De~Decker, Marcus Magnor, Philippe Bekaert, Edilson de Aguiar, Naveed Ahmed, Christian Theobalt, and Anita Sellent.
\newblock Floating textures.
\newblock \emph{Computer Graphics Forum (Proc. of Eurographics {EG})}, 27\penalty0 (2):\penalty0 409--418, 2008.
\newblock Received the Best Student Paper Award at Eurographics 2008.

\bibitem[Habermann et~al.(2020)Habermann, Xu, Zollhoefer, Pons-Moll, and Theobalt]{habermann20}
Marc Habermann, Weipeng Xu, Michael Zollhoefer, Gerard Pons-Moll, and Christian Theobalt.
\newblock Deepcap: Monocular human performance capture using weak supervision.
\newblock \emph{Proceedings of the Conference on Computer Vision and Pattern Recognition (CVPR)}, 1:\penalty0 1, 2020.

\bibitem[Habermann et~al.(2021{\natexlab{a}})Habermann, Liu, Xu, Zollhoefer, Pons-Moll, and Theobalt]{habermann2021real}
Marc Habermann, Lingjie Liu, Weipeng Xu, Michael Zollhoefer, Gerard Pons-Moll, and Christian Theobalt.
\newblock Real-time deep dynamic characters.
\newblock \emph{ACM Transactions on Graphics (TOG)}, 40\penalty0 (4):\penalty0 1--16, 2021{\natexlab{a}}.

\bibitem[Habermann et~al.(2021{\natexlab{b}})Habermann, Xu, Zollhoefer, Pons-Moll, and Theobalt]{habermann21a}
Marc Habermann, Weipeng Xu, Michael Zollhoefer, Gerard Pons-Moll, and Christian Theobalt.
\newblock A deeper look into deepcap.
\newblock In \emph{IEEE Transactions on Pattern Analysis and Machine Intelligence (TPAMI)}, pages 1--1. IEEE, 2021{\natexlab{b}}.

\bibitem[Habermann et~al.(2023)Habermann, Liu, Xu, Pons-Moll, Zollhoefer, and Theobalt]{habermann2022hdhumans}
Marc Habermann, Lingjie Liu, Weipeng Xu, Gerard Pons-Moll, Michael Zollhoefer, and Christian Theobalt.
\newblock Hdhumans: A hybrid approach for high-fidelity digital humans.
\newblock \emph{Proc. ACM Comput. Graph. Interact. Tech.}, 6\penalty0 (3), 2023.

\bibitem[Heusel et~al.(2017)Heusel, Ramsauer, Unterthiner, Nessler, and Hochreiter]{heusel17}
Martin Heusel, Hubert Ramsauer, Thomas Unterthiner, Bernhard Nessler, and Sepp Hochreiter.
\newblock Gans trained by a two time-scale update rule converge to a local nash equilibrium.
\newblock In \emph{Proceedings of the 31st International Conference on Neural Information Processing Systems}, page 6629–6640, Red Hook, NY, USA, 2017. Curran Associates Inc.

\bibitem[I{\c{s} }{\i}k et~al.(2023)I{\c{s} }{\i}k, Rünz, Georgopoulos, Khakhulin, Starck, Agapito, and Nie{\ss}ner]{HumanRF2023}
Mustafa I{\c{s} }{\i}k, Martin Rünz, Markos Georgopoulos, Taras Khakhulin, Jonathan Starck, Lourdes Agapito, and Matthias Nie{\ss}ner.
\newblock {HumanRF}: High-fidelity neural radiance fields for humans in motion.
\newblock \emph{{ACM} Transactions on Graphics}, 42\penalty0 (4):\penalty0 1--12, 2023.

\bibitem[Jiang et~al.(2022)Jiang, Yi, Samei, Tuzel, and Ranjan]{jiang2022neuman}
Wei Jiang, Kwang~Moo Yi, Golnoosh Samei, Oncel Tuzel, and Anurag Ranjan.
\newblock Neuman: Neural human radiance field from a single video.
\newblock In \emph{European Conference on Computer Vision (ECCV)}, 2022.

\bibitem[Kappel et~al.(2021)Kappel, Golyanik, Elgharib, Henningson, Seidel, Castillo, Theobalt, and Magnor]{kappel2020high-fidelity}
Moritz Kappel, Vladislav Golyanik, Mohamed Elgharib, Jann-Ole Henningson, Hans-Peter Seidel, Susana Castillo, Christian Theobalt, and Marcus Magnor.
\newblock High-fidelity neural human motion transfer from monocular video.
\newblock In \emph{Computer Vision and Pattern Recognition (CVPR)}, pages 1541--1550, 2021.

\bibitem[Kavan et~al.(2007{\natexlab{a}})Kavan, Collins, \v{Z}\'{a}ra, and O'Sullivan]{dualquaternion2007}
Ladislav Kavan, Steven Collins, Ji\v{r}\'{\i} \v{Z}\'{a}ra, and Carol O'Sullivan.
\newblock Skinning with dual quaternions.
\newblock In \emph{Proceedings of the 2007 Symposium on Interactive 3D Graphics and Games}, page 39–46, New York, NY, USA, 2007{\natexlab{a}}. Association for Computing Machinery.

\bibitem[Kavan et~al.(2007{\natexlab{b}})Kavan, Collins, {\v{Z}}{\'a}ra, and O'Sullivan]{kavan2007skinning}
Ladislav Kavan, Steven Collins, Ji{\v{r}}{\'\i} {\v{Z}}{\'a}ra, and Carol O'Sullivan.
\newblock Skinning with dual quaternions.
\newblock In \emph{Proceedings of the 2007 symposium on Interactive 3D graphics and games}, pages 39--46. ACM, 2007{\natexlab{b}}.

\bibitem[Kingma and Ba(2014)]{kingma2017adam}
Diederik Kingma and Jimmy Ba.
\newblock Adam: A method for stochastic optimization.
\newblock \emph{International Conference on Learning Representations}, 2014.

\bibitem[Kwon et~al.(2023{\natexlab{a}})Kwon, Kim, Ceylan, and Fuchs]{kwon2023neuralimage}
Youngjoong Kwon, Dahun Kim, Duygu Ceylan, and Henry Fuchs.
\newblock Neural image-based avatars: Generalizable radiance fields for human avatar modeling.
\newblock In \emph{The Eleventh International Conference on Learning Representations, {ICLR} 2023, Kigali, Rwanda, May 1-5, 2023}. OpenReview.net, 2023{\natexlab{a}}.

\bibitem[Kwon et~al.(2023{\natexlab{b}})Kwon, Liu, Fuchs, Habermann, and Theobalt]{kwon2023deliffas}
Youngjoong Kwon, Lingjie Liu, Henry Fuchs, Marc Habermann, and Christian Theobalt.
\newblock Deliffas: Deformable light fields for fast avatar synthesis.
\newblock \emph{Advances in Neural Information Processing Systems}, 2023{\natexlab{b}}.

\bibitem[Li et~al.(2022{\natexlab{a}})Li, Tanke, Vo, Zollhofer, Gall, Kanazawa, and Lassner]{li2022tava}
Ruilong Li, Julian Tanke, Minh Vo, Michael Zollhofer, Jurgen Gall, Angjoo Kanazawa, and Christoph Lassner.
\newblock Tava: Template-free animatable volumetric actors.
\newblock In \emph{European Conference on Computer Vision (ECCV)}, 2022{\natexlab{a}}.

\bibitem[Li et~al.(2022{\natexlab{b}})Li, Slavcheva, Zollh\"ofer, Green, Lassner, Kim, Schmidt, Lovegrove, Goesele, Newcombe, and Lv]{Li2022CVPR}
Tianye Li, Mira Slavcheva, Michael Zollh\"ofer, Simon Green, Christoph Lassner, Changil Kim, Tanner Schmidt, Steven Lovegrove, Michael Goesele, Richard Newcombe, and Zhaoyang Lv.
\newblock Neural 3d video synthesis from multi-view video.
\newblock In \emph{Computer Vision and Pattern Recognition (CVPR)}, 2022{\natexlab{b}}.

\bibitem[Lim et~al.(2017)Lim, Son, Kim, Nah, and Lee]{lim2017enhanced}
Bee Lim, Sanghyun Son, Heewon Kim, Seungjun Nah, and Kyoung~Mu Lee.
\newblock Enhanced deep residual networks for single image super-resolution, 2017.

\bibitem[Lin et~al.(2022)Lin, Peng, Xu, Yan, Shuai, Bao, and Zhou]{lin2022efficient}
Haotong Lin, Sida Peng, Zhen Xu, Yunzhi Yan, Qing Shuai, Hujun Bao, and Xiaowei Zhou.
\newblock Efficient neural radiance fields with learned depth-guided sampling.
\newblock In \emph{SIGGRAPH Asia Conference Proceedings}, 2022.

\bibitem[Lin et~al.(2021)Lin, Ryabtsev, Sengupta, Curless, Seitz, and Kemelmacher-Shlizerman]{lin2021real}
Shanchuan Lin, Andrey Ryabtsev, Soumyadip Sengupta, Brian~L Curless, Steven~M Seitz, and Ira Kemelmacher-Shlizerman.
\newblock Real-time high-resolution background matting.
\newblock In \emph{Conference on Computer Vision and Pattern Recognition (CVPR)}, pages 8762--8771, 2021.

\bibitem[Liu et~al.(2019)Liu, Xu, Zollh\"{o}fer, Kim, Bernard, Habermann, Wang, and Theobalt]{liu19}
Lingjie Liu, Weipeng Xu, Michael Zollh\"{o}fer, Hyeongwoo Kim, Florian Bernard, Marc Habermann, Wenping Wang, and Christian Theobalt.
\newblock Neural rendering and reenactment of human actor videos.
\newblock \emph{ACM Transactions on Graphics (TOG)}, 38\penalty0 (5), 2019.

\bibitem[Liu et~al.(2020)Liu, Xu, Habermann, Zollhöfer, Bernard, Kim, Wang, and Theobalt]{liu20}
Lingjie Liu, Weipeng Xu, Marc Habermann, Michael Zollhöfer, Florian Bernard, Hyeongwoo Kim, Wenping Wang, and Christian Theobalt.
\newblock Neural human video rendering by learning dynamic textures and rendering-to-video translation.
\newblock \emph{Transactions on Visualization and Computer Graphics (TVCG)}, PP:\penalty0 1--1, 2020.

\bibitem[Liu et~al.(2021)Liu, Habermann, Rudnev, Sarkar, Gu, and Theobalt]{liu2021neural}
Lingjie Liu, Marc Habermann, Viktor Rudnev, Kripasindhu Sarkar, Jiatao Gu, and Christian Theobalt.
\newblock Neural actor: Neural free-view synthesis of human actors with pose control.
\newblock \emph{ACM Trans. Graph.}, 40\penalty0 (6), 2021.

\bibitem[Liu et~al.(2023)Liu, Gao, Meuleman, Tseng, Saraf, Kim, Chuang, Kopf, and Huang]{liu2023robust}
Yu-Lun Liu, Chen Gao, Andreas Meuleman, Hung-Yu Tseng, Ayush Saraf, Changil Kim, Yung-Yu Chuang, Johannes Kopf, and Jia-Bin Huang.
\newblock Robust dynamic radiance fields.
\newblock In \emph{Computer Vision and Pattern Recognition (CVPR)}, 2023.

\bibitem[Mildenhall et~al.(2019)Mildenhall, Srinivasan, Ortiz-Cayon, Kalantari, Ramamoorthi, Ng, and Kar]{mildenhall2019local}
Ben Mildenhall, Pratul~P Srinivasan, Rodrigo Ortiz-Cayon, Nima~Khademi Kalantari, Ravi Ramamoorthi, Ren Ng, and Abhishek Kar.
\newblock Local light field fusion: Practical view synthesis with prescriptive sampling guidelines.
\newblock \emph{ACM Transactions on Graphics (TOG)}, 38\penalty0 (4):\penalty0 1--14, 2019.

\bibitem[Noguchi et~al.(2021)Noguchi, Sun, Lin, and Harada]{2021narf}
Atsuhiro Noguchi, Xiao Sun, Stephen Lin, and Tatsuya Harada.
\newblock Neural articulated radiance field.
\newblock In \emph{International Conference on Computer Vision}, 2021.

\bibitem[Pang et~al.(2023)Pang, Zhu, Kortylewski, Theobalt, and Habermann]{zhu2023ash}
Haokai Pang, Heming Zhu, Adam Kortylewski, Christian Theobalt, and Marc Habermann.
\newblock Ash: Animatable gaussian splats for efficient and photoreal human rendering.
\newblock 2023.

\bibitem[Pavlakos et~al.(2019)Pavlakos, Choutas, Ghorbani, Bolkart, Osman, Tzionas, and Black]{smplx}
Georgios Pavlakos, Vasileios Choutas, Nima Ghorbani, Timo Bolkart, Ahmed A.~A. Osman, Dimitrios Tzionas, and Michael~J. Black.
\newblock Expressive body capture: 3d hands, face, and body from a single image.
\newblock In \emph{{IEEE} Conference on Computer Vision and Pattern Recognition, {CVPR} 2019, Long Beach, CA, USA, June 16-20, 2019}, pages 10975--10985. Computer Vision Foundation / {IEEE}, 2019.

\bibitem[Peng et~al.(2021{\natexlab{a}})Peng, Dong, Wang, Zhang, Shuai, Bao, and Zhou]{peng2021animatable}
Sida Peng, Junting Dong, Qianqian Wang, Shangzhan Zhang, Qing Shuai, Hujun Bao, and Xiaowei Zhou.
\newblock Animatable neural radiance fields for human body modeling.
\newblock \emph{ICCV}, 2021{\natexlab{a}}.

\bibitem[Peng et~al.(2021{\natexlab{b}})Peng, Zhang, Xu, Wang, Shuai, Bao, and Zhou]{peng2020neural}
Sida Peng, Yuanqing Zhang, Yinghao Xu, Qianqian Wang, Qing Shuai, Hujun Bao, and Xiaowei Zhou.
\newblock Neural body: Implicit neural representations with structured latent codes for novel view synthesis of dynamic humans.
\newblock \emph{CVPR}, 1\penalty0 (1):\penalty0 9054--9063, 2021{\natexlab{b}}.

\bibitem[Raj et~al.(2021)Raj, Tanke, Hays, Vo, Stoll, and Lassner]{Raj_ANR}
Amit Raj, Julian Tanke, James Hays, Minh Vo, Carsten Stoll, and Christoph Lassner.
\newblock Anr: Articulated neural rendering for virtual avatars.
\newblock In \emph{IEEE/CVF Conference on Computer Vision and Pattern Recognition (CVPR)}, 2021.

\bibitem[Remelli et~al.(2022)Remelli, Bagautdinov, Saito, Wu, Simon, Wei, Guo, Cao, Prada, Saragih, et~al.]{remelli2022drivable}
Edoardo Remelli, Timur Bagautdinov, Shunsuke Saito, Chenglei Wu, Tomas Simon, Shih-En Wei, Kaiwen Guo, Zhe Cao, Fabian Prada, Jason Saragih, et~al.
\newblock Drivable volumetric avatars using texel-aligned features.
\newblock In \emph{ACM SIGGRAPH 2022 Conference Proceedings}, pages 1--9, 2022.

\bibitem[Ronneberger et~al.(2015)Ronneberger, Fischer, and Brox]{ronneberger2015unet}
Olaf Ronneberger, Philipp Fischer, and Thomas Brox.
\newblock U-net: Convolutional networks for biomedical image segmentation.
\newblock In \emph{Medical Image Computing and Computer-Assisted Intervention - {MICCAI} 2015 - 18th International Conference Munich, Germany, October 5 - 9, 2015, Proceedings, Part {III}}, pages 234--241. Springer, 2015.

\bibitem[Shysheya et~al.(2019)Shysheya, Zakharov, Aliev, Bashirov, Burkov, Iskakov, Ivakhnenko, Malkov, Pasechnik, Ulyanov, Vakhitov, and Lempitsky]{Shysheya_2019_CVPR}
Aliaksandra Shysheya, Egor Zakharov, Kara-Ali Aliev, Renat Bashirov, Egor Burkov, Karim Iskakov, Aleksei Ivakhnenko, Yury Malkov, Igor Pasechnik, Dmitry Ulyanov, Alexander Vakhitov, and Victor Lempitsky.
\newblock Textured neural avatars.
\newblock In \emph{The IEEE Conference on Computer Vision and Pattern Recognition (CVPR)}, 2019.

\bibitem[Siarohin et~al.(2019)Siarohin, Lathuilière, Tulyakov, Ricci, and Sebe]{Siarohin_2019_NeurIPS}
Aliaksandr Siarohin, Stéphane Lathuilière, Sergey Tulyakov, Elisa Ricci, and Nicu Sebe.
\newblock First order motion model for image animation.
\newblock In \emph{Conference on Neural Information Processing Systems (NeurIPS)}, 2019.

\bibitem[Sorkine and Alexa(2007)]{sorkine2007rigid}
Olga Sorkine and Marc Alexa.
\newblock As-rigid-as-possible surface modeling.
\newblock In \emph{Proceedings of the Fifth Eurographics Symposium on Geometry Processing}. Eurographics Association, 2007.

\bibitem[Sumner et~al.(2007)Sumner, Schmid, and Pauly]{sumner2007embedded}
Robert~W. Sumner, Johannes Schmid, and Mark Pauly.
\newblock Embedded deformation for shape manipulation.
\newblock \emph{ACM Trans. Graph.}, 26\penalty0 (3), 2007.

\bibitem[{TheCaptury}(2020)]{captury}
{TheCaptury}.
\newblock {The Captury}.
\newblock \url{http://www.thecaptury.com/}, 2020.

\bibitem[{Treedys}(2020)]{treedys2020}
{Treedys}.
\newblock {Treedys}.
\newblock \url{https://www.treedys.com/}, 2020.

\bibitem[Tretschk et~al.(2021)Tretschk, Tewari, Golyanik, Zollh\"{o}fer, Lassner, and Theobalt]{tretschk2020nonrigid}
Edgar Tretschk, Ayush Tewari, Vladislav Golyanik, Michael Zollh\"{o}fer, Christoph Lassner, and Christian Theobalt.
\newblock Non-rigid neural radiance fields: Reconstruction and novel view synthesis of a dynamic scene from monocular video.
\newblock In \emph{{IEEE} International Conference on Computer Vision ({ICCV})}. {IEEE}, 2021.

\bibitem[Tung et~al.(2009)Tung, Nobuhara, and Matsuyama]{tung09}
T. Tung, S. Nobuhara, and T. Matsuyama.
\newblock Complete multi-view reconstruction of dynamic scenes from probabilistic fusion of narrow and wide baseline stereo.
\newblock In \emph{International Conference on Computer Vision (ICCV)}, pages 1709--1716. IEEE, 2009.

\bibitem[Vlasic et~al.(2009)Vlasic, Peers, Baran, Debevec, Popovi{\'c}, Rusinkiewicz, and Matusik]{vlasic09}
Daniel Vlasic, Pieter Peers, Ilya Baran, Paul Debevec, Jovan Popovi{\'c}, Szymon Rusinkiewicz, and Wojciech Matusik.
\newblock Dynamic shape capture using multi-view photometric stereo.
\newblock \emph{ACM Transactions on Graphics (TOG)}, 28\penalty0 (5):\penalty0 174, 2009.

\bibitem[Wang et~al.(2023{\natexlab{a}})Wang, MacDonald, Jeni, and Lucey]{Wang2023FSDNeRF}
Chaoyang Wang, Lachlan~Ewen MacDonald, L\'aszl\'o~A. Jeni, and Simon Lucey.
\newblock Flow supervision for deformable nerf.
\newblock In \emph{Computer Vision and Pattern Recognition (CVPR)}, 2023{\natexlab{a}}.

\bibitem[Wang et~al.(2021)Wang, Wang, Genova, Srinivasan, Zhou, Barron, Martin-Brualla, Snavely, and Funkhouser]{wang2021ibrnet}
Qianqian Wang, Zhicheng Wang, Kyle Genova, Pratul Srinivasan, Howard Zhou, Jonathan~T. Barron, Ricardo Martin-Brualla, Noah Snavely, and Thomas Funkhouser.
\newblock Ibrnet: Learning multi-view image-based rendering.
\newblock In \emph{CVPR}, 2021.

\bibitem[Wang et~al.(2022{\natexlab{a}})Wang, Schwarz, Geiger, and Tang]{ARAH:ECCV:2022}
Shaofei Wang, Katja Schwarz, Andreas Geiger, and Siyu Tang.
\newblock Arah: Animatable volume rendering of articulated human sdfs.
\newblock In \emph{European Conference on Computer Vision}, 2022{\natexlab{a}}.

\bibitem[Wang et~al.(2023{\natexlab{b}})Wang, Han, Habermann, Daniilidis, Theobalt, and Liu]{wang2022neus2}
Yiming Wang, Qin Han, Marc Habermann, Kostas Daniilidis, Christian Theobalt, and Lingjie Liu.
\newblock Neus2: Fast learning of neural implicit surfaces for multi-view reconstruction.
\newblock In \emph{International Conference on Computer Vision (ICCV)}, pages 3295--3306, 2023{\natexlab{b}}.

\bibitem[Wang et~al.(2022{\natexlab{b}})Wang, Nam, Stuyck, Lombardi, Zollhoefer, Hodgins, and Lassner]{wang2022hvh}
Ziyan Wang, Giljoo Nam, Tuur Stuyck, Stephen Lombardi, Michael Zollhoefer, Jessica Hodgins, and Christoph Lassner.
\newblock Hvh: Learning a hybrid neural volumetric representation for dynamic hair performance capture.
\newblock In \emph{IEEE/CVF Conference on Computer Vision and Pattern Recognition (CVPR)}, 2022{\natexlab{b}}.

\bibitem[Waschbüsch et~al.(2005)Waschbüsch, Würmlin, Cotting, Sadlo, and Gross]{pointbased}
Michael Waschbüsch, Stephan Würmlin, Daniel Cotting, Filip Sadlo, and Markus Gross.
\newblock Scalable 3d video of dynamic scenes.
\newblock \emph{The Visual Computer}, 21:\penalty0 629--638, 2005.

\bibitem[Weng et~al.(2022)Weng, Curless, Srinivasan, Barron, and Kemelmacher-Shlizerman]{weng2022humannerf}
Chung-Yi Weng, Brian Curless, Pratul~P Srinivasan, Jonathan~T Barron, and Ira Kemelmacher-Shlizerman.
\newblock Humannerf: Free-viewpoint rendering of moving people from monocular video.
\newblock In \emph{Computer Vision and Pattern Recognition (CVPR)}, pages 16210--16220, 2022.

\bibitem[Xiang et~al.(2022)Xiang, Bagautdinov, Stuyck, Prada, Romero, Xu, Saito, Guo, Smith, Shiratori, Sheikh, Hodgins, and Wu]{donglai2022dressing}
Donglai Xiang, Timur Bagautdinov, Tuur Stuyck, Fabian Prada, Javier Romero, Weipeng Xu, Shunsuke Saito, Jingfan Guo, Breannan Smith, Takaaki Shiratori, Yaser Sheikh, Jessica Hodgins, and Chenglei Wu.
\newblock Dressing avatars: Deep photorealistic appearance for physically simulated clothing.
\newblock \emph{ACM Trans. Graph.}, 41\penalty0 (6), 2022.

\bibitem[Xu et~al.(2011)Xu, Liu, Stoll, Tompkin, Bharaj, Dai, Seidel, Kautz, and Theobalt]{Xu:SIGGRPAH:2011}
Feng Xu, Yebin Liu, Carsten Stoll, James Tompkin, Gaurav Bharaj, Qionghai Dai, Hans-Peter Seidel, Jan Kautz, and Christian Theobalt.
\newblock Video-based characters: Creating new human performances from a multi-view video database.
\newblock In \emph{ACM SIGGRAPH 2011 Papers}, pages 32:1--32:10, New York, NY, USA, 2011. ACM.

\bibitem[Xu et~al.(2021)Xu, Alldieck, and Sminchisescu]{xu2021hnerf}
Hongyi Xu, Thiemo Alldieck, and Cristian Sminchisescu.
\newblock H-nerf: Neural radiance fields for rendering and temporal reconstruction of humans in motion.
\newblock In \emph{Neural Information Processing Systems}, 2021.

\bibitem[Yu et~al.(2021)Yu, Ye, Tancik, and Kanazawa]{yu2020pixelnerf}
Alex Yu, Vickie Ye, Matthew Tancik, and Angjoo Kanazawa.
\newblock {pixelNeRF}: Neural radiance fields from one or few images.
\newblock In \emph{CVPR}, 2021.

\bibitem[Yuksel(2015)]{poissondisk2015}
Cem Yuksel.
\newblock Sample elimination for generating poisson disk sample sets.
\newblock \emph{Computer Graphics Forum}, 34, 2015.

\bibitem[Zhang et~al.(2021)Zhang, Liu, Ye, Zhao, Zhang, Wu, Zhang, Xu, and Yu]{zhang2021stnerf}
Jiakai Zhang, Xinhang Liu, Xinyi Ye, Fuqiang Zhao, Yanshun Zhang, Minye Wu, Yingliang Zhang, Lan Xu, and Jingyi Yu.
\newblock Editable free-viewpoint video using a layered neural representation.
\newblock \emph{ACM Transactions on Graphics (TOG)}, 40\penalty0 (4):\penalty0 1--18, 2021.

\bibitem[Zhang et~al.(2018)Zhang, Isola, Efros, Shechtman, and Wang]{zhang2018unreasonable}
R. Zhang, P. Isola, A.~A. Efros, E. Shechtman, and O. Wang.
\newblock The unreasonable effectiveness of deep features as a perceptual metric.
\newblock In \emph{Conference on Computer Vision and Pattern Recognition (CVPR)}, pages 586--595, Los Alamitos, CA, USA, 2018. IEEE Computer Society.

\bibitem[Zhao et~al.(2022)Zhao, Jiang, Yao, Zhang, Wang, Dai, Zhong, Zhang, Wu, Xu, and Yu]{zhao2022humanp}
Fuqiang Zhao, Yuheng Jiang, Kaixin Yao, Jiakai Zhang, Liao Wang, Haizhao Dai, Yuhui Zhong, Yingliang Zhang, Minye Wu, Lan Xu, and Jingyi Yu.
\newblock Human performance modeling and rendering via neural animated mesh.
\newblock \emph{ACM Trans. Graph.}, 41\penalty0 (6), 2022.

\end{thebibliography}
}
\clearpage
\appendix

\section{Supplementary Material}
%
%
We first present more technical details on our character model (Sec.~\ref{subsec:ddc_supp}), 
and the depth testing approach for texel visibility (Sec.~\ref{subsec:depth_testing}).
Then, we provide more implementation as well as data processing details (Sec.~\ref{subsec:impl_det} and Sec.~\ref{subsec:data}), additional results, and applications of our method (Sec.~\ref{subsec:add_results}, Sec.~\ref{subsec:demo}).
We further show additional comparisons (Sec.~\ref{subsec:add_comp}), ablations (Sec.~\ref{subsec:add_ablations}), and analysis of methods closest to our setting (Sec.~\ref{subsec:enerf_dva}). Finally, we talk about the limitations of our approach (Sec.~\ref{subsec:future}).

\subsection{Deformable Character Model} \label{subsec:ddc_supp}

Our character model takes a temporal motion $M= \{(\boldsymbol{\theta}_{t-W},\boldsymbol{\alpha}_{t-W},\boldsymbol{z}_{t-W})....(\boldsymbol{\theta}_{t},\boldsymbol{\alpha}_{t},\boldsymbol{z}_{t})\}$  as input and
deforms a template mesh capable of modelling loose clothing. Here $\boldsymbol{\theta}_{t} \in \mathbb{R}^{P},\boldsymbol{\alpha}_t \in \mathbb{R}^{3}, \boldsymbol{z}_t \in \mathbb{R}^{3}$ refer to the skeleton degrees of freedom, root translation, and root rotation, respectively. We leverage the explicit character representation of \citet{habermann2021real}
\begin{equation}
C_i (\boldsymbol{\theta}_t, \boldsymbol{\alpha}_t, \boldsymbol{z}_t, \boldsymbol{A},\boldsymbol{T}, \boldsymbol{d}_i) = \boldsymbol{v}_i,
\end{equation}
as it
is differentiable, real-time, and models loose clothing. 
In their character formulation, the initial template $T$ is downsampled to an embedded graph~\cite{sumner2007embedded} $G$ with $K$ nodes, and the parameters $\boldsymbol{A} \in \mathbb{R}^{K \times 3},\boldsymbol{T} \in \mathbb{R}^{K \times 3}$ are the rotation and translation of each of the $K$ nodes stacked on top of each other, describing the coarse deformation of $T$ in canonical space. $\boldsymbol{d}_i \in  \mathbb{R}^{3}$ is the per-vertex displacement in canonical space.
The final location 
   \begin{equation}
\boldsymbol{y}_i = \boldsymbol{d}_i+ \sum_{k \in N_{\mathrm{vn}(i)}}w_{i,k}R(\boldsymbol{a}_k)(\hat{\boldsymbol{v}}_i-\boldsymbol{g}_k)+\boldsymbol{g}_k+\boldsymbol{t}_k
   \end{equation}
of a vertex in canonical space $\boldsymbol{y}_i \in \mathbb{R}^{3}$ is determined by the weighted addition of the rotation and translation of its neighbours in the embedded graph $N_{\mathrm{vn}(i)}$, and finally adding the per-vertex deformation $\boldsymbol{d}_i$. 
Here, $w_{i,k}$ is the weight the $i$th vertex assigns to node k. $R(\cdot)$ is the function that converts Euler angles to matrices. $\boldsymbol{a}_k$, $\boldsymbol{t}_k$ are $k$th rows of $\boldsymbol{A}$, $\boldsymbol{T}$ respectively. $\hat{\boldsymbol{v}}_i$ is the $i$th vertex of $T$ and $\boldsymbol{g}_k$ is the $k$th node of G.

Their model predicts $\boldsymbol{A}$, $\boldsymbol{T}$, and $\boldsymbol{d}_i$ using structure-aware graph neural networks\cite{habermann2021real}, referred to as $f_{\mathrm{eg}}(e(M))$ and  $f_{\mathrm{delta}}(d(M))$. $e(\cdot)$ and $d(\cdot)$ are their proposed motion to embedded graph embedding and motion to vertex embedding. Finally, they apply the skeleton pose to the deformed vertex in canonical space $\boldsymbol{y_i}$ to obtain the final deformed vertex location 
\begin{equation}
    \boldsymbol{v}_i=\boldsymbol{z}+\sum_{k \in N_{\mathrm{vn(i)}}}w_{i,k}(R_{\mathrm{sk},k}(\boldsymbol{\theta,\alpha})\boldsymbol{y}_i+\boldsymbol{t}_{\mathrm{sk},k}(\boldsymbol{\theta,\alpha})),
\end{equation}
where the rotation $R_{\mathrm{sk},k}$, and translation $\boldsymbol{t}_{\mathrm{sk},k}$ for node $k$ are determined by Dual Quaternion Skinning~\cite{kavan2007skinning}. The vertex matrix $\boldsymbol{V} \in \mathbb{R}^{N \times 3}$ can be obtained by stacking $\boldsymbol{v_i}$.
%
%
\subsection{Depth Testing}
\label{subsec:depth_testing}
We address the texel visibility problem by using a depth testing approach.
We assign a 3D depth to each texel $T_{\mathrm{pos}} \in \mathbb{R}^{TW \times TH \times 3}$, using the barycentric coordinates of the face it lands on. 
Then, we compute the depth $D_{c,V} \in \mathbb{R}^{ H \times W \times 3 }$ and the texel to pixel mapping $  F_{\mathrm{warp_{c,V}}} : (u,v) \rightarrow (x,y) $ for a particular view $c$ using differentiable rasterisation from DeepCap~\cite{habermann20,habermann21a} given the camera parameters and the deformed vertex positions $\boldsymbol{V}$.
If the depth at a particular texel $(u,v)$ is within a $\epsilon$ norm ball to the depth of the pixel it lands on, we mark it as visible, i.e.
\small
\begin{equation}
T_{\mathrm{vis},i}(u,v)=|D_{i,V}(F_{\mathrm{warp_{i,V}}}(u,v))-T_{\mathrm{pos}}(u,v)| < \epsilon.
\end {equation}
\normalsize

%
%
\begin{table}
\small
\begin{center}
\begin{tabular}{@{}c|cc|cc@{}}
    \toprule
     &  \multicolumn{2}{c|}{Novel View} & \multicolumn{2}{c}{Novel Pose} \\
     \midrule
                                \textbf{Method}     & \textbf{CD}$\downarrow$   & \textbf{HD}$\downarrow$  & \textbf{CD}$\downarrow$   & \textbf{HD}$\downarrow$  \\
    \midrule
    DDC	                          &    11.53         &    11.24              &    15.5  &    15.6      \\ 
    HDHumans	                 &    11.52          &    11.21              &    13.7  &    13.5    \\
   
    \textbf{Ours}	       &   \textbf{10.4}                &  \textbf{9.3}  &    \textbf{13.1}  &    \textbf{12.9}     \\  
    \bottomrule
\end{tabular}
 \vspace{-8pt}
\caption
{
\textbf{Quantitative Comparison on Surface Distance.} 
We evaluate the surface tracking in the \textit{novel view} and \textit{novel pose} setting on subject \textit{S1} and report the Chamfer (CD) and Hausdorff (HD) distance with respect to the ground truth. Note that our approach outperforms previous works, thanks to the additional Chamfer penalty. This is significant in enhancing the quality of our projective texturing pipeline. Error reported in $mm$. 
}
\label{tab:comparison_chamfer}	
\end{center}
\end{table}
%
%
%
%
\begin{table}
\centering
\small
\begin{tabular}{@{}c|cc@{}}
    \toprule
    \textbf{Component} & \textbf{FPS}$\uparrow$  & \textbf{Latency}$\downarrow$ \\
    \midrule

    Character Model	 &   100               &  0.010$s$           \\ 
   Projective Texturing	 &    27               &  0.037$s$         \\ 
    TexfeatNet	 &   100               &  0.010$s$        \\ 
    SRNet	 &   31.25               &  0.032$s$       \\ 
    \bottomrule
\end{tabular}
\caption{
\textbf{Runtime Breakdown.}
Here, we present a component-wise runtime breakdown of our pipeline in terms of frames per second (fps) and latency (in seconds $s$). Note that all of our components run within the real-time limit of 25 fps. 
}
\label{tab:component_wise_breakdown}	
\end{table}
%
%
\subsection{Additional Data Processing Details}\label{subsec:data}
Each actor in our dataset is scanned using a commercially available 3D scanner \cite{treedys2020} where the mesh is obtained from multi-view stereo reconstruction\footnote{\url{https://www.agisoft.com}}. Following this, we downsample the high-resolution mesh to around $9000$ faces for each character. For character rigging, we apply markerless motion capture~\cite{captury} on the multi-view images from the scanner to obtain the skeletal pose. Given the pose and the template scan, we apply Blender's\footnote{\url{https://www.blender.org}} automated skinning weight computation. 
The UV parameterization is obtained from a photometric stereo reconstruction software. However, the effect of UV parameterization and optimizing the UV parameterization is a future work that we believe merits further investigation.
\subsection{Implementation Details}\label{subsec:impl_det}
The obtained multi-view frames are processed with foreground segmentation~\cite{lin2021real} and per-frame mesh reconstruction using NeuS2~\cite{wang2022neus2}.
Motion tracking is obtained with a commercial markerless capture system~\cite{captury}.
Our method is implemented in TensorFlow~\cite{tensorflow} and trained using the Adam optimizer~\cite{kingma2017adam} with a constant learning rate of $10^{-4}$ until convergence. 

The character model and the TexFeatNet are supervised on images with a resolution of $1028\times{752}$ pixels, and the SRNet is trained with a full image resolution of $4112\times{3008}$ pixels.
The projective texturing module uses images of resolution $4112\times{3008}$ as input, and the texture maps are generated at a resolution of $1024\times{1024}$ pixels.
We randomly sample 10K points from the reconstructed surface for the Chamfer loss for every frame. 

TexFeatNet utilizes a UNet architecture \cite{ronneberger2015unet}, and SRNet is implemented as a shallow architecture with enhanced residual blocks \cite{lim2017enhanced}.
Our complete framework is trained using two Nvidia A100 GPUs with 80GB memory and a batch size of four.
Tab.~\ref{tab:component_wise_breakdown} provides a runtime breakdown of our modules.

\begin{figure}
\centering
  \includegraphics[width=0.9\linewidth]{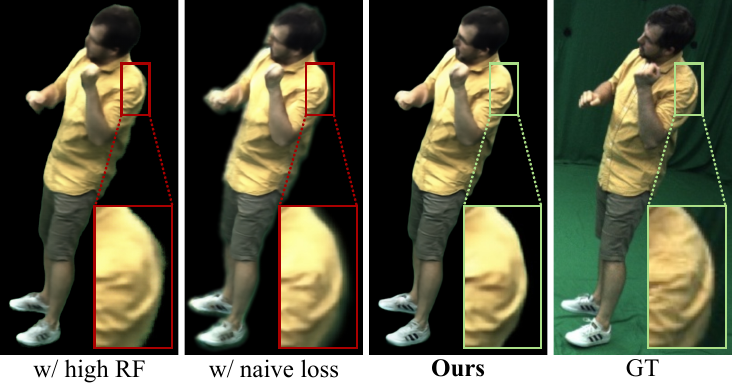}
  \caption{\textbf{Qualitative Ablations.} Note that our design choices for the SRNet module, lead to qualitatively better results, especially at the borders of the human. }
  \label{fig:12_add_ablt}
\end{figure}

\begin{figure}
\centering
  \includegraphics[width=0.9\linewidth]{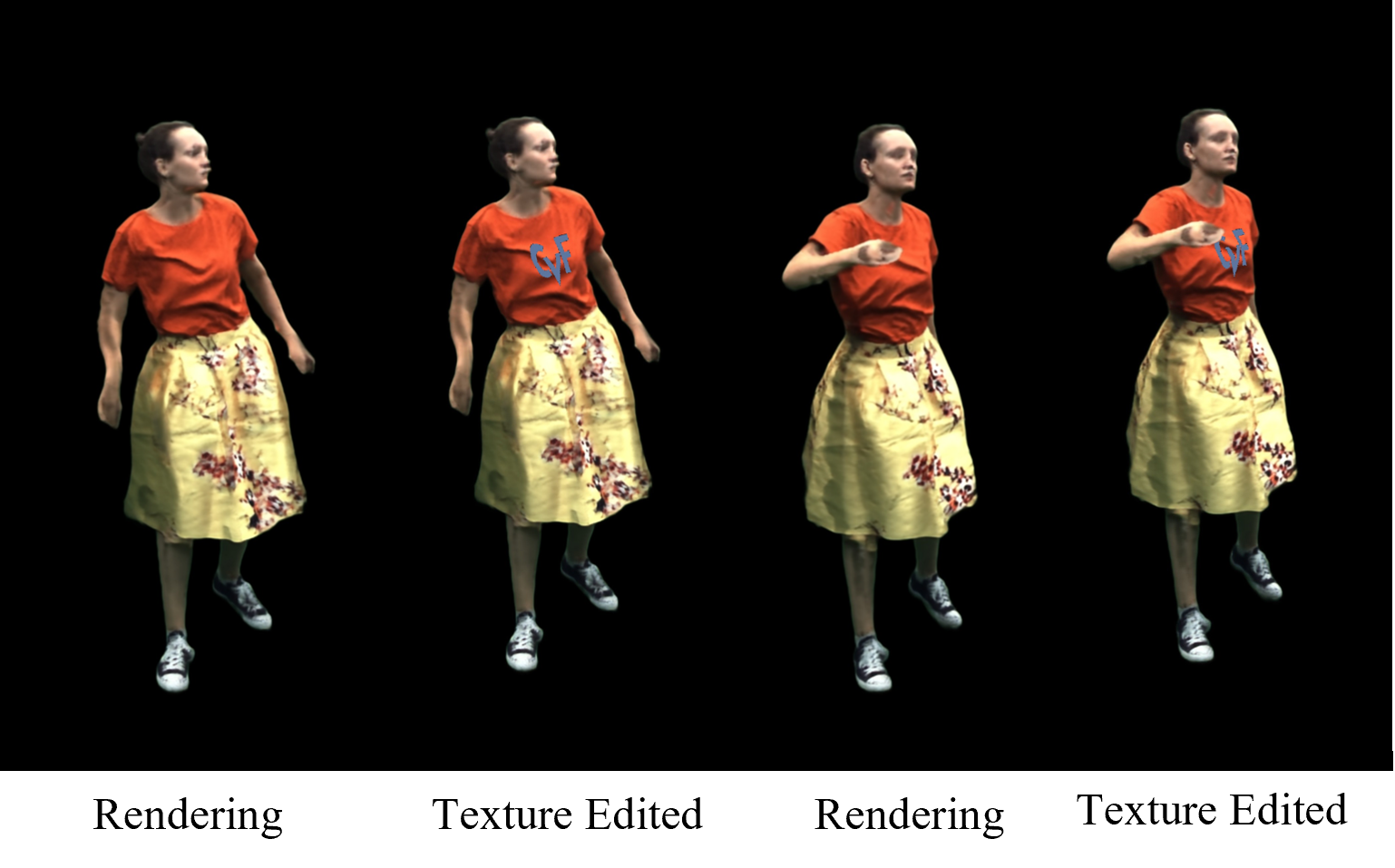}
  \caption{\textbf{Application: Texture Editing.} As we use explicit textures as an underlying latent for appearance, and they are temporally and spatially aligned, we can perform 2D texture edits such as adding a logo onto the character's shirt.}
  \label{fig:13_texture_editing}
\end{figure}
%
%
\begin{table}
\centering
\small
\begin{tabular}{@{}ccccc@{}}
    \toprule
    \textbf{Method} & \textbf{PSNR}$\uparrow$  & \textbf{LPIPS}$\downarrow$ \footnotesize{($\times 1000$)} & \textbf{FID}$\downarrow$ & \textbf{Res.} \\
    \midrule

    w/ Naive $\mathcal{L}_\mathrm{sr}$	 &   27.01               &  55.10        &   37.25          &   4K    \\ 
    w/ High RF	 &    28.13               &  36.18        &   19.97          &   4K    \\ 
    \textbf{Ours}	 &   28.75                &  \textbf{32.4}   &   \textbf{17.42}        &   4K     \\ 
    \bottomrule
\end{tabular}
\caption{
\textbf{Quantitative Ablations.}
Here, we ablate some design choices in our SRNet module, in the novel pose setting on subject \textit{S1}. Note that choosing our $\mathcal{L}_\mathrm{sr}$ formulation, and keeping a shallower architecture lead to better performance. 
}
\label{tab:ablation_add}	
\end{table}
%
%
%
%
\begin{table}
\centering
\small
\begin{tabular}{@{}ccccc@{}}
    \toprule
    \textbf{Method} & \textbf{PSNR}$\uparrow$  & \textbf{LPIPS}$\downarrow$ \footnotesize{($\times 1000$)} & \textbf{FID}$\downarrow$ & \textbf{Res.} \\
    \midrule

    w/o Texture 	 &   25.82               &  41.36        &   55.17          &   1K    \\ 
    w/o Features	 &    28.37               &  31.30        &   21.05         &   1K    \\ 
     w/o Chamfer	 &    27.83               &  30.35        &   15.76         &   1K    \\ 
    w/o SR	 &    28.42               &  31.11      &   20.85         &   1K    \\ 
    w/o 4K	 &    \textbf{28.85}               &  30.50      &   18.01         &   1K    \\ 
    \textbf{Ours}	 &   28.03                &  \textbf{28.49}   &   \textbf{13.26}        &   1K     \\ 
    \midrule
    Ours w/o 4K   &  \textbf{27.72}         &   34.49         &  20.89  &  4K    \\ 
    \textbf{Ours}	 &   27.22               &  \textbf{33.17}   &   \textbf{15.26}        &   4K     \\ 
    \bottomrule
\end{tabular}
\caption{
\textbf{Quantitative Ablations.}
Here, we ablate some major design choices in the novel pose setting on subject \textit{S2} (loose clothing). Note that the efficacy of our design choices translates along subjects. 
}
\label{tab:ablation_fran}	
\end{table}
%
%
\begin{figure}
\centering
    \includegraphics[width=1.\linewidth]{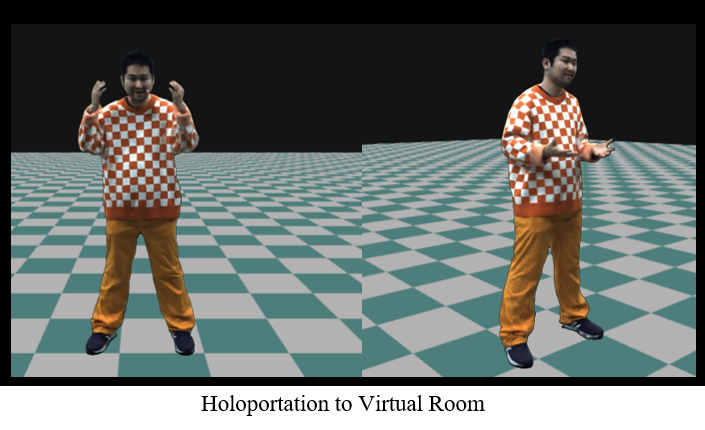}
    \vspace{-24pt}
    \caption{
    \textbf{Application: Holoportation.} Our method produces high-quality expressions, hands, and wrinkles in real time; hence, it is well suited for telepresence applications like placing a character in a virtual room. }
    \label{fig:holoported_room}
\end{figure}
\subsection{Additional Results}\label{subsec:add_results}
\begin{figure*}
\centering
  \includegraphics[width=0.9\textwidth]{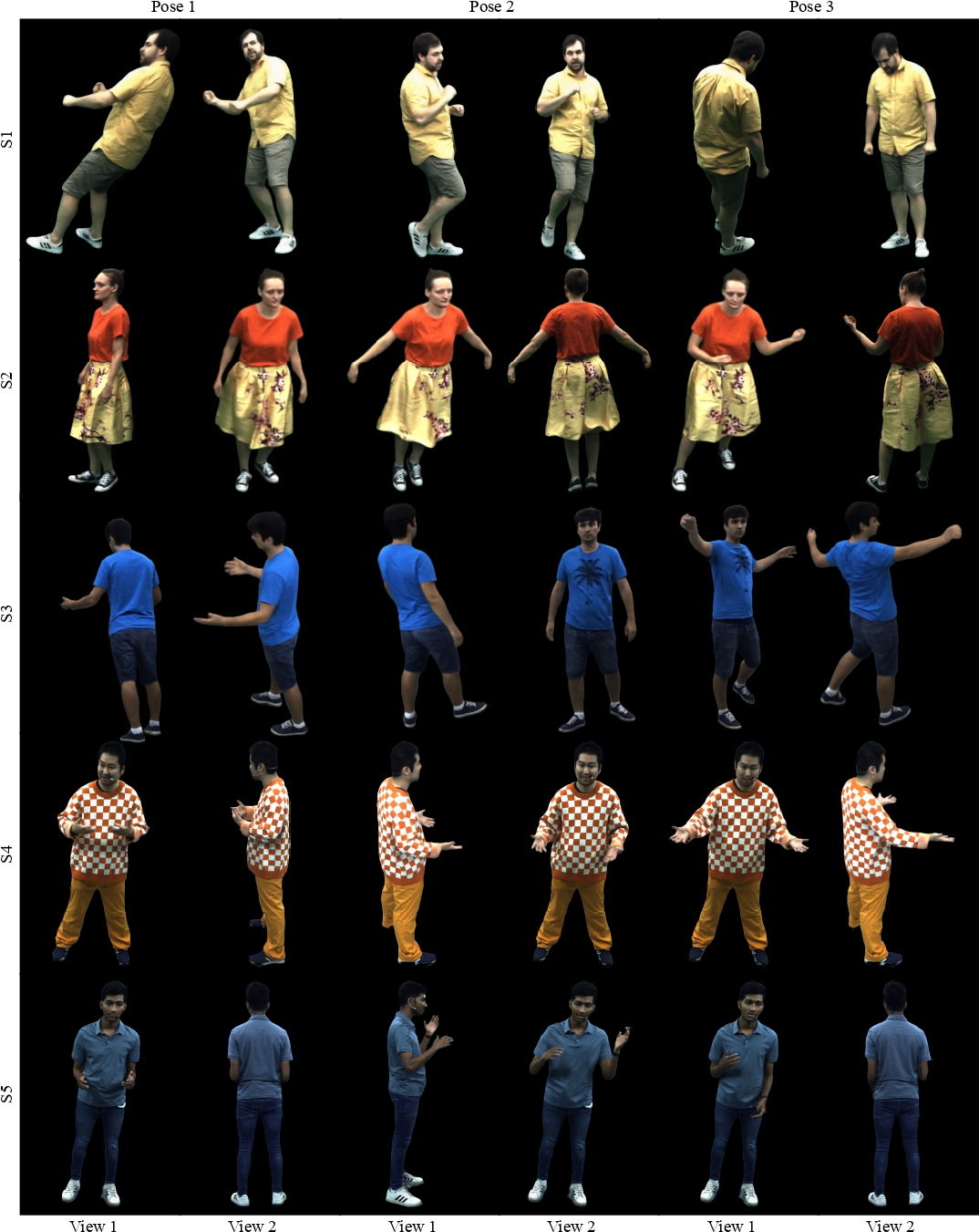}
  \caption{\textbf{Results from our method in the \textit{novel view synthesis} setting for five different subjects}. Our method faithfully reconstructs fine details such as wrinkles (S1, S3), loose clothes with large deformations (S2), rich textures (S4), and hand poses.}
  \label{fig:6_novel_view}
\end{figure*}

\begin{figure*}
\centering
  \includegraphics[width=0.9\textwidth]{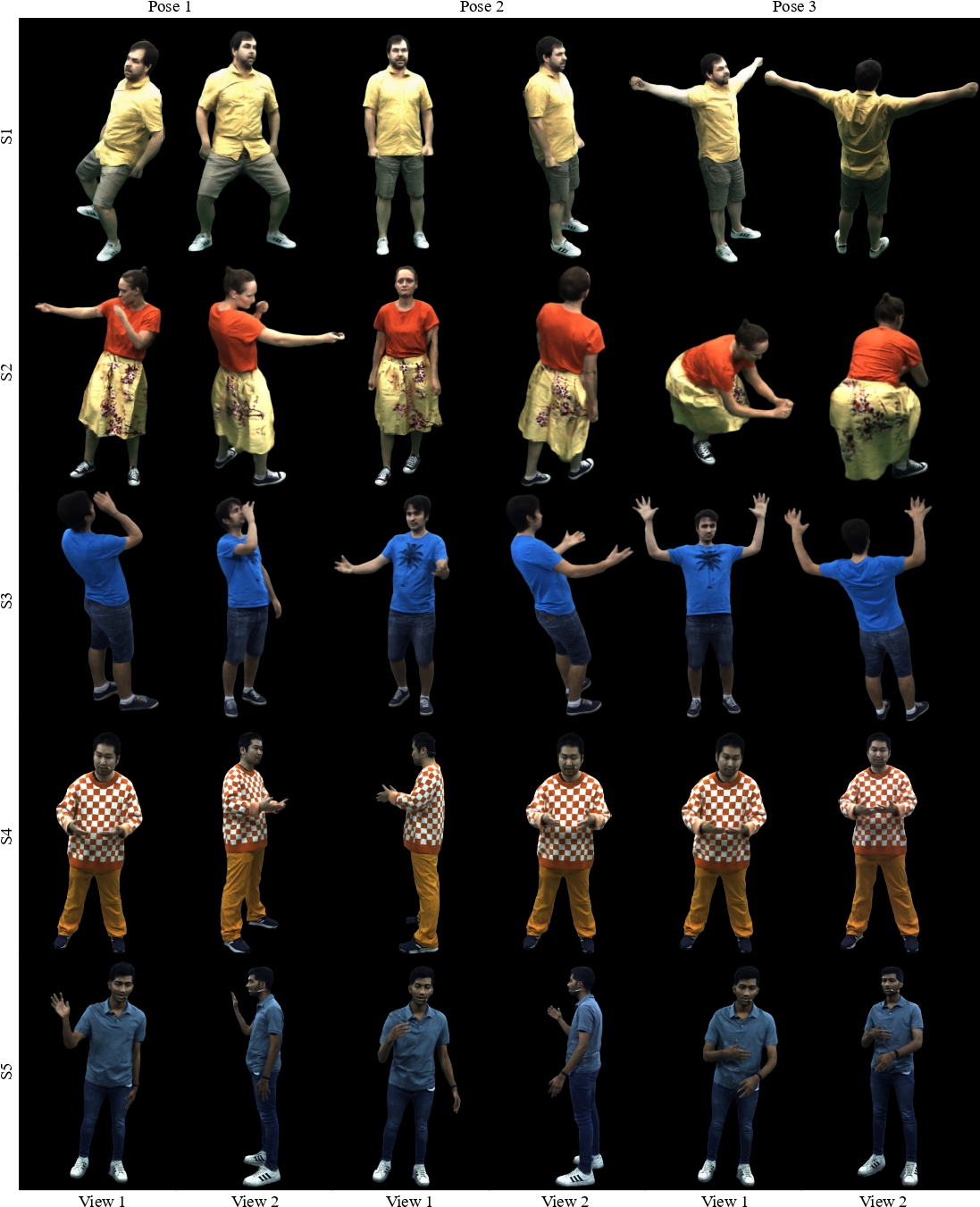}
  \caption{\textbf{Results from our method in the \textit{novel pose synthesis} setting for five different subjects}. The results are rendered from a novel view point not seen during training. Our method results in impressive renderings, providing very realistic wrinkle patterns and high-frequency details.}
  \label{fig:7_novel_pose}
\end{figure*}

We also provide additional results for our method on more subjects in the novel view and novel pose setting (Fig. \ref{fig:6_novel_view} and Fig. \ref{fig:7_novel_pose}).
Our method also allows for exciting applications like texture editing (Fig. \ref{fig:13_texture_editing}) and placing the character in a virtual room (Fig. \ref{fig:holoported_room}).

\subsection{End-to-end Sparse Camera Demo}\label{subsec:demo}
\begin{figure}
\centering
    \includegraphics[width=1.\linewidth]{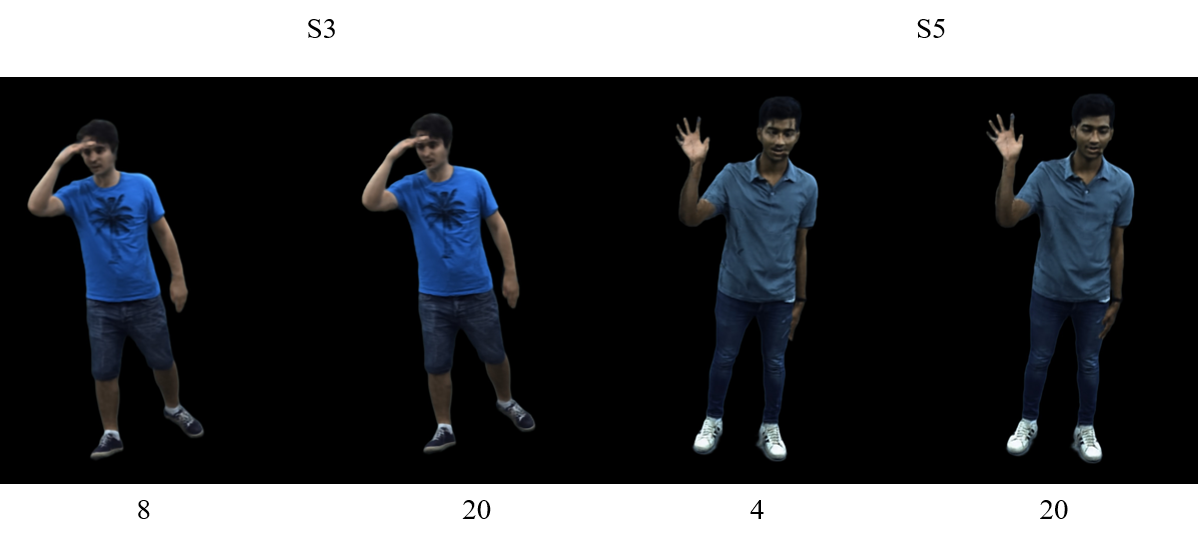}
    \vspace{-24pt}
    \caption{
    \textbf{Sparse Camera Pose Tracking.}
    We present results using 3D pose tracking from fewer cameras (numbers below image represent number of cameras used). The tracking inaccuracies lead to artifacts; however, the quality remains relatively high even with four cameras.
    }
    \label{fig:less_cameras}
\end{figure}

We additionally show the result of our method using a 3D skeletal pose recovered from four and eight cameras, respectively, for different subjects (Fig. \ref{fig:less_cameras}). 
Note that the result with four cameras demonstrates that our method can also be integrated into an end-to-end sparse camera setup.

\section{Additional Comparisons}
\label{subsec:add_comp}
\begin{figure}
    %

        \includegraphics[width=1.\linewidth]{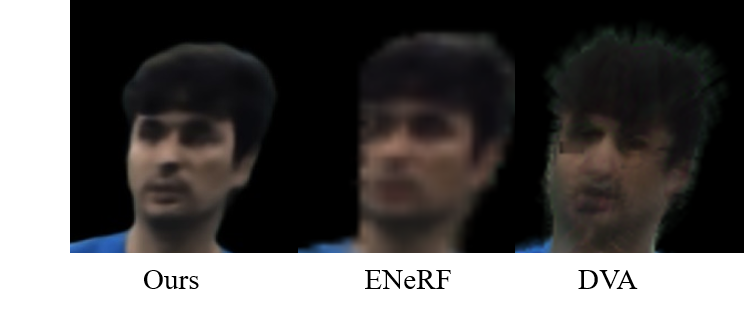}
   
    \caption{
\textbf{Comparison with a zoomed-in virtual view.} Here we present a result, where we render a view close to the face. Notice the ability of our method to capture facial expressions with much higher fidelity. 
    }
     \label{fig:face_comparison}
\end{figure}

\begin{figure*}
\centering
  \includegraphics[width=0.9\textwidth]{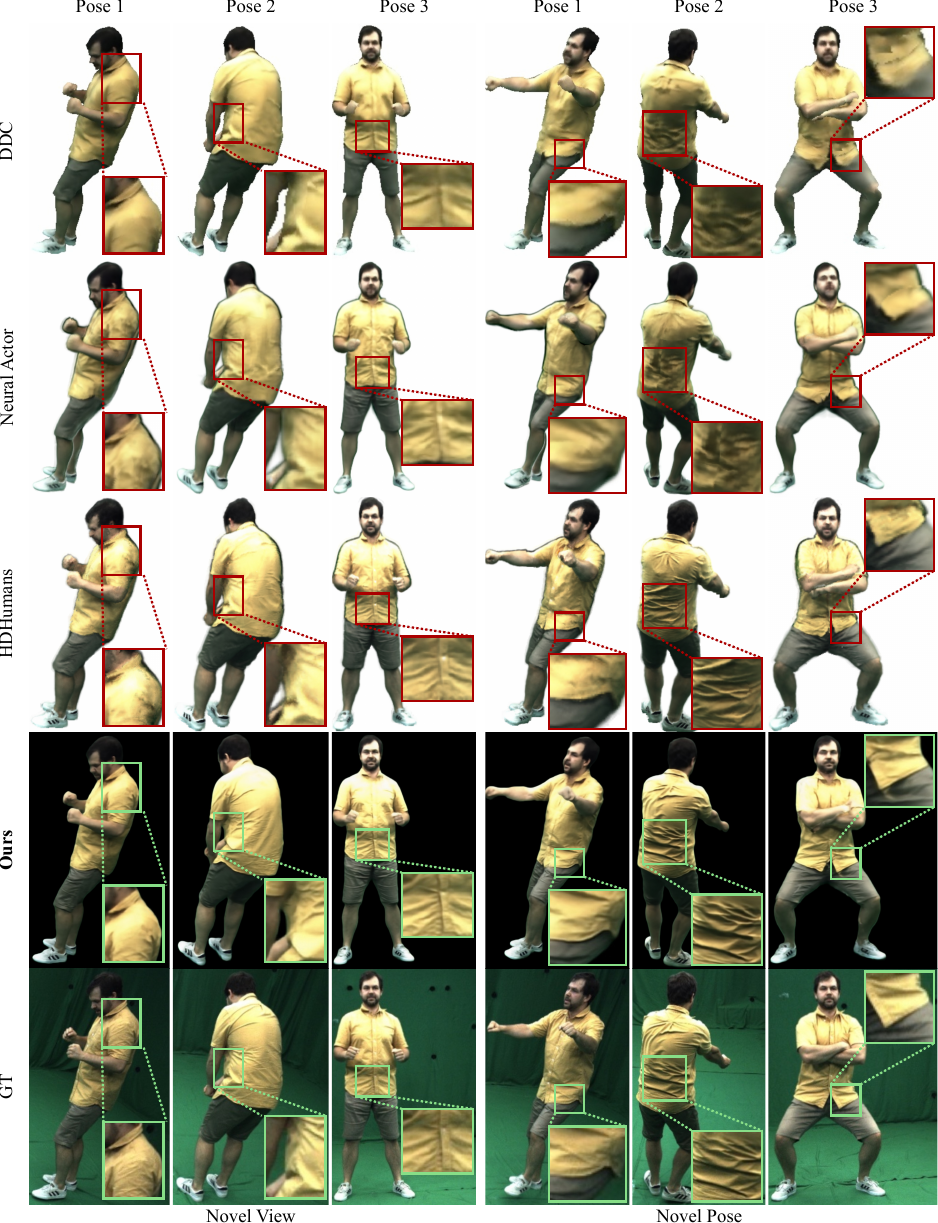}
  \caption{\textbf{Comparison of our method with previous pose-driven approaches}. Note how DDC~\cite{habermann2021real} and Neural Actor (NA)~\cite{liu2021neural} fail to produce high-frequency details and cloth wrinkles, while our method produces high-quality renderings for both novel view and novel pose settings. }
  \label{fig:8_comparison_pose_sota}
\end{figure*}

We present additional qualitative comparisons in the novel view and novel pose setting to animatable representations (Fig.~\ref{fig:8_comparison_pose_sota}) and real-time sparse image-driven methods (Fig. \ref{fig:9_comparison_enerf} and Fig.~\ref{fig:10_comparison_enerf_franzi}). We also present a zoomed-in face comparison of our method to show our efficacy in capturing face expressions (Fig.~\ref{fig:face_comparison}).
In Tab.~\ref{tab:comparison_chamfer}, we compare the quality of our geometry reconstruction against competing methods. 
Our approach provides quantitative improvements in the Chamfer (CD) and Hausdorff metrics (HD). 
\subsection{Additional Ablations}
\label{subsec:add_ablations}
Here, we provide some additional ablation results (Fig. \ref{fig:12_add_ablt} and Tab. \ref{tab:ablation_add}). 
We ablate our SRNet module by replacing it with a naive loss that is the same as $\mathcal{L}_\mathrm{ren} $ and find that this performs quantitatively worse and also qualitatively (especially at the borders). 
Also, replacing a shallow SRNet architecture with a deeper architecture that utilizes UNet \cite{ronneberger2015unet}, and additional upsampling layers leads to worse multi-view consistency, which can be seen quantitatively and qualitatively. 
Additionally, we also add an ablation for our major components, on a subject in loose clothing (Tab.~\ref{tab:ablation_fran}).
\begin{figure}
\centering
    \includegraphics[width=1.\linewidth]{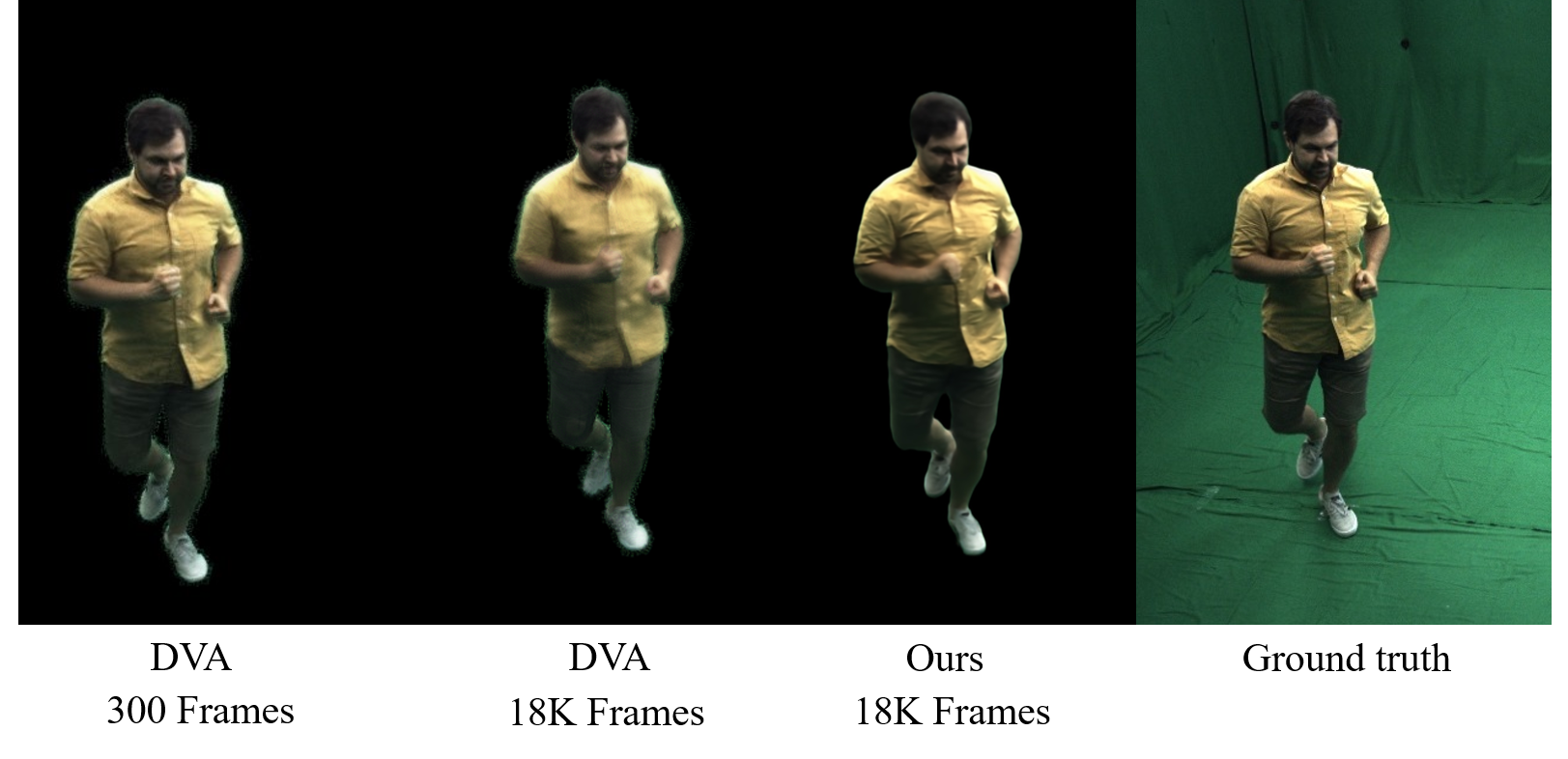}
    \vspace{-24pt}
    \caption{
    \textbf{Large-scale Training.}
  We present a novel view synthesis (replay), the result from DVA when it is trained on 300 frames, and when it's trained on 18000 frames. Note that the frame visualized is part of the training frames. While DVA can capture high-fidelity details, when trained with fewer frames, the performance deteriorates as the number of frames increases. In contrast, ours maintains high fidelity, even as the number of frames increases.
    }
    \label{fig:dva_train}
\end{figure}
\subsection{ENeRF and DVA Analysis}\label{subsec:enerf_dva}
\begin{figure*}[ht]
\centering
  \includegraphics[width=0.9\textwidth]{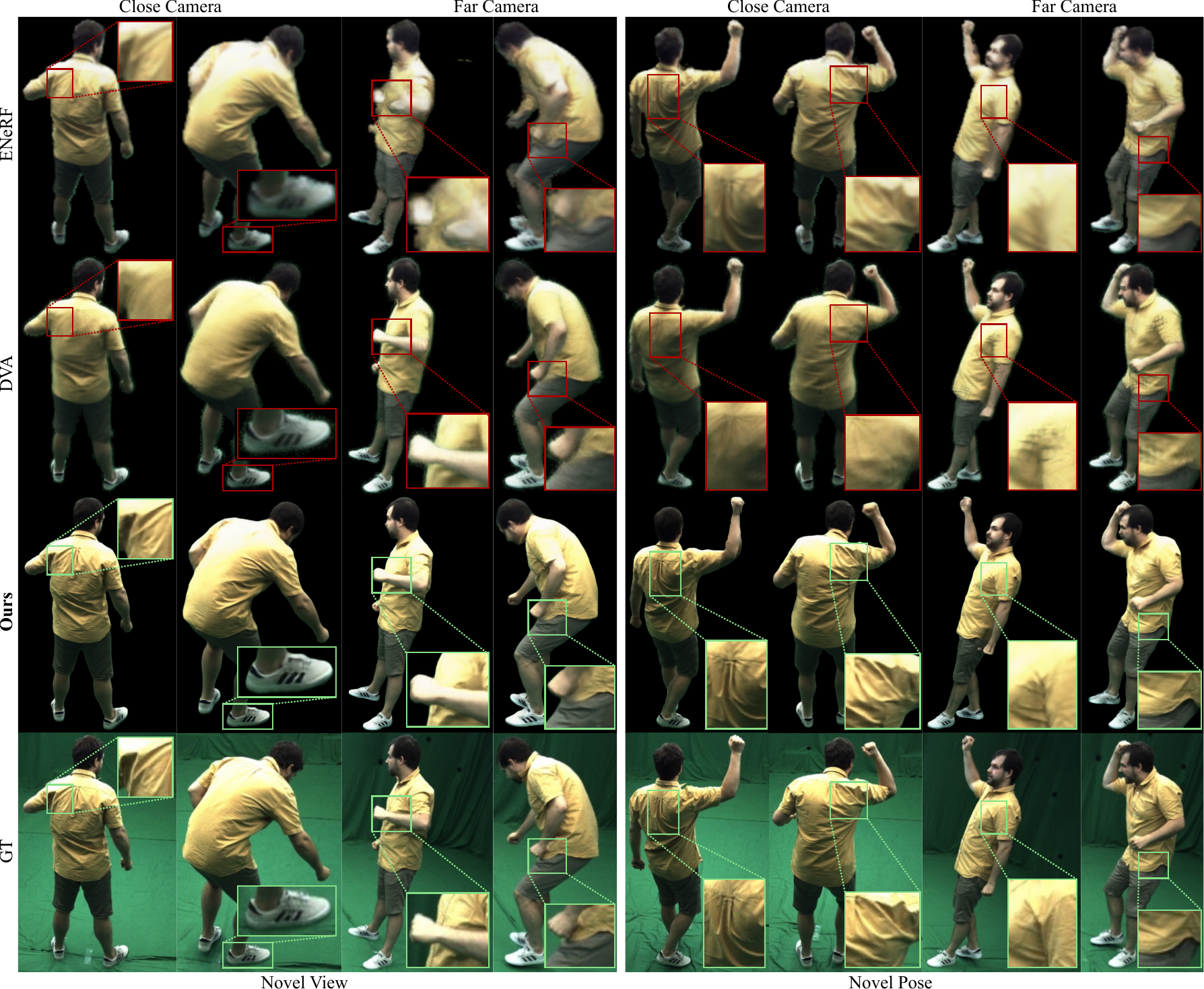}
  \caption{
  \textbf{Comparison of our method with existing real-time approaches that take images as input}. Note how ENeRF~\cite{lin2022efficient} produces artifacts under novel camera views and how DVA~\cite{remelli2022drivable} suffers from inaccurate tracking, resulting in blurry renderings. On the contrary, our method generalizes well to far camera viewpoints and produces sharp results.}
  \label{fig:9_comparison_enerf}
  
\end{figure*}

\begin{figure}[tb]
\centering
  \includegraphics[width=0.47\textwidth]{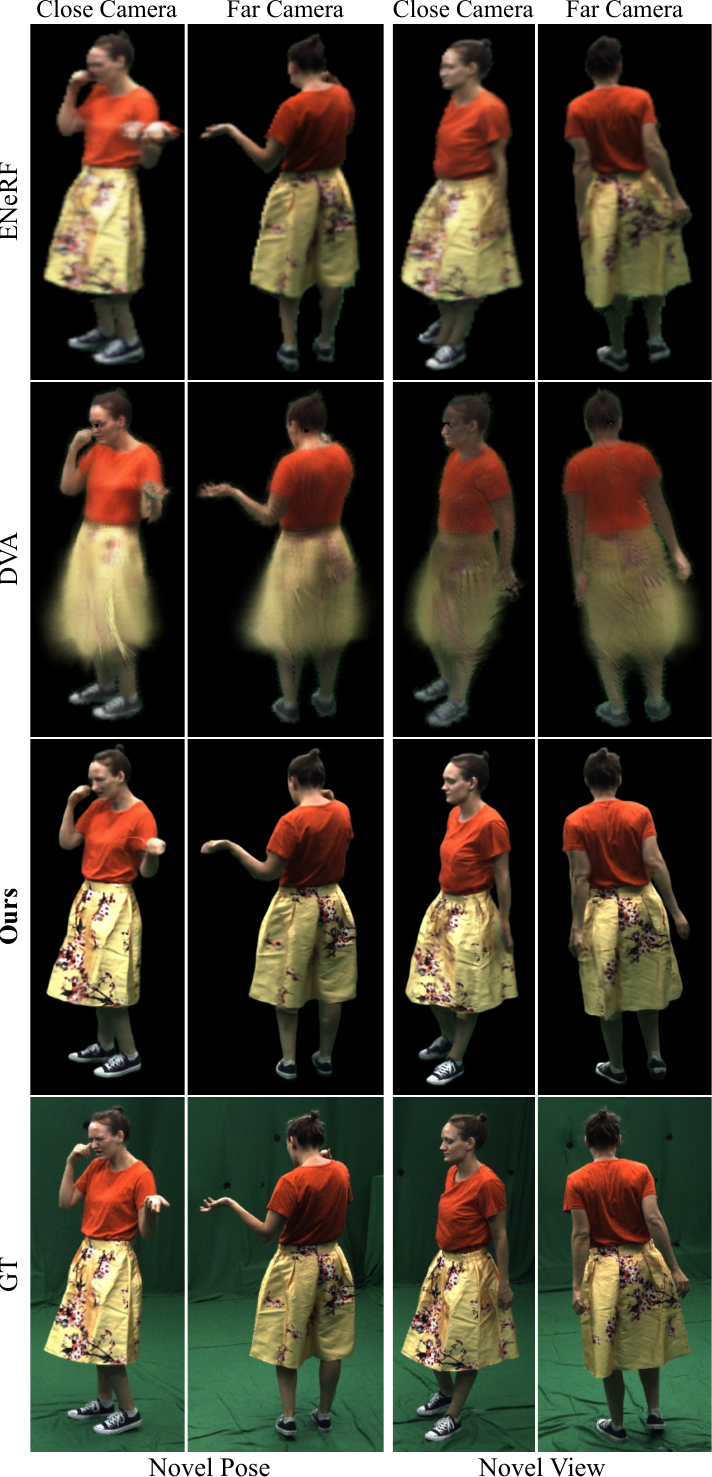}
  \caption{\textbf{Comparison to ENeRF and DVA  on  Loose Clothing}. ENeRF~\cite{lin2022efficient} can handle loose clothing reasonably well, but suffers from artifacts in other body parts (like in the hands). DVA~\cite{remelli2022drivable} fails to converge on loose-clothed subjects, leading to artifacts in areas like the skirt. In contrast, as our character model can handle loose clothing, we produce sharp results with details preserved.}
  \label{fig:10_comparison_enerf_franzi}
\end{figure}

We additionally provide more details on how we compared to ENeRF \cite{lin2022efficient} and DVA \cite{remelli2022drivable}, as they are closely related to our setting. 
ENeRF is a general method that generates novel views from sparse source camera views and even demonstrates impressive generalizability to completely unseen scenes.
 However, in their setting, they utilize \textit{nearest} views \textit{at test time} from a dense setup to produce the target view, which violates our requirement for only a fixed number of cameras at test time.
Hence, to compare with them, we retrain their method in the same setting as ours (inference using four \textit{fixed cameras}). 
We observe their method faces multi-view consistency artifacts due to a lack of reliable human priors for depth, and as their final color is a weighted combination of the source colors (see Fig.~\ref{fig:9_comparison_enerf} and Fig.~\ref{fig:10_comparison_enerf_franzi}).

DVA achieves photorealistic telepresence in real-time from sparse cameras and 3D skeletal pose. 
They released a version of their source code, which relies on SMPLX \cite{smplx} for mesh tracking and provided scripts to reproduce results on the ZJU dataset \cite{peng2020neural}.
First, we reproduce results on the ZJU dataset. Then, to test their robustness to loose clothing and challenging poses, we evaluate on the DynaCap dataset. 
We used SMPLX tracking from twenty cameras and trained their method in the same setting as ours. We observe while they can do replay and novel view synthesis with high quality if we train only on small sequences (300 frames), their result becomes blurry when trained on more frames (see Fig. \ref{fig:dva_train}). 
We hypothesize that this is caused by the fact that their volume primitives have to model fine-scale deformations and appearance at the same time, while the capacity of the network is too limited to model both of those aspects of the human.
Additionally, we observe that the model fails to converge on loose clothing, as their volume regularizer prevents primitives from moving far away from the SMPLX initialization, which is imperative in the case of loose clothing (see Fig.~\ref{fig:10_comparison_enerf_franzi}). Ours, in contrast, deals with deformations separately in the explicit character model, which allows us to maintain appearance quality even as the number of training frames increases.
\par 

\subsection{Limitations and Future Work} \label{subsec:future}
Though our work is a clear step towards more immersive and photorealistic avatars, there are remaining challenges yet to be addressed in the future. 
For example, our method, similar to other prior works, does not allow to model topological changes, e.g. opening a jacket. 
Future work could explore layered human representations, potentially able to model such effects. 
Flickering artifacts occur in our method, due to inconsistent color calibration of the multi-view cameras. 
We believe a joint optimization of camera parameters, i.e. color, extrinsic, and intrinsic calibration, can potentially resolve this limitation. 
Additionally, tracking errors in the case of fast motions, e.g.~jumping, may result in artifacts in the renderings.
Tightly entangling the tracking and rendering might resolve this in the future.
Moreover, we currently require a dense studio setup to acquire the photoreal avatar. 
In the future, we plan to explore more lightweight setups even for training the model.
Last, our projective texturing takes the learned geometry from the deformable character model as input while not being able to further refine it throughout the training. 
We believe differentiable projective texturing could be an interesting direction to tackle this problem.
\begin{figure*}
\centering
  \includegraphics[width=1.\textwidth]{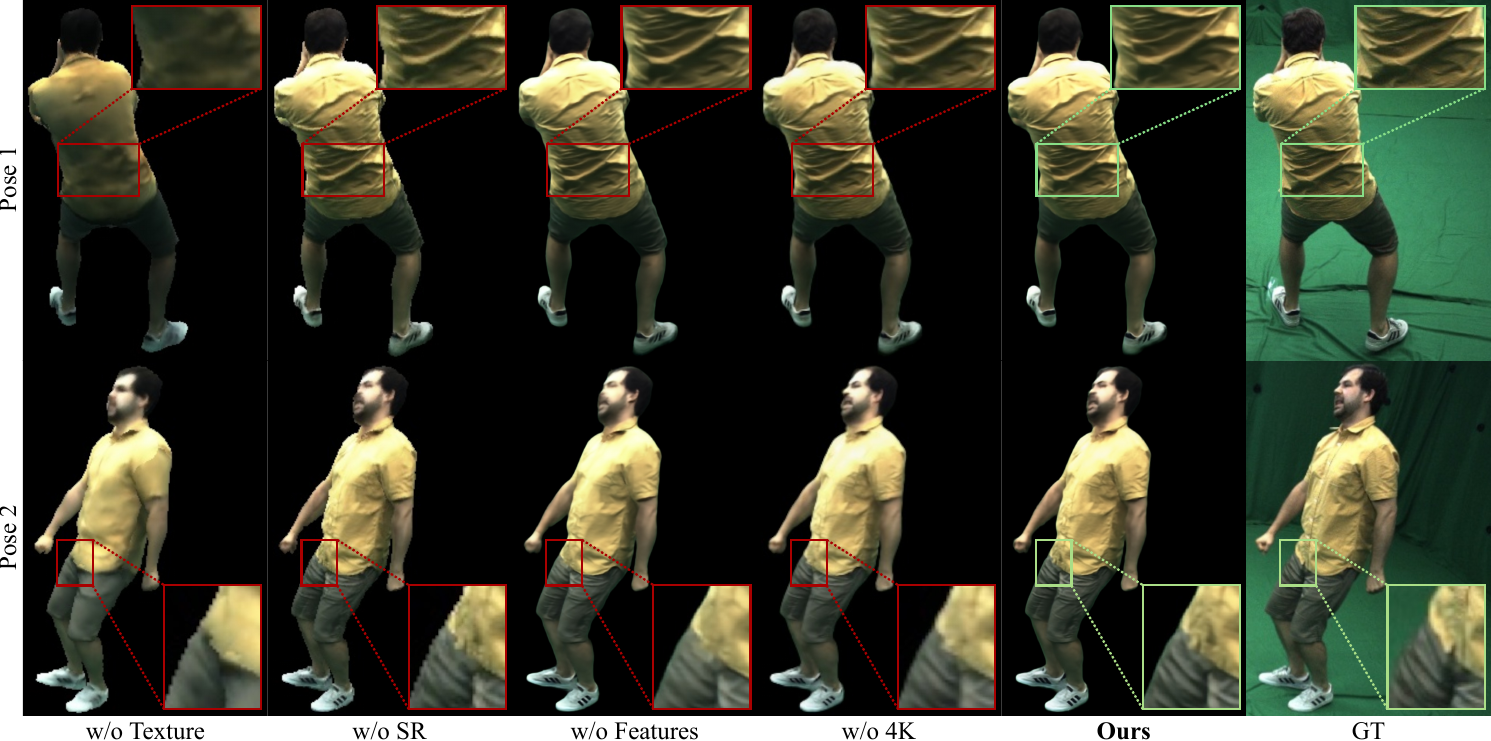}
  \caption{Qualitative results considering the contributions of the components from our method. Without texture, results fail to reproduce fine details. Without the SR module and the features as additional input further degrades the borders and the high-frequency details. Finally, without 4K training, the results are less sharp compared to our method.}
  \label{fig:11_ablation}
\end{figure*}

\newpage

\end{document}